\documentclass[11pt]{article}

\usepackage[preprint]{acl}

\usepackage{times}
\usepackage{latexsym}

\usepackage[T1]{fontenc}

\usepackage[utf8]{inputenc}

\usepackage{microtype}

\usepackage{inconsolata}

\usepackage{graphicx}

\usepackage{amsmath}
\usepackage{pifont}
\usepackage{booktabs}
\usepackage{multirow}
\title{HeteroSpec: Leveraging Contextual Heterogeneity for Efficient Speculative Decoding}

\author{
Siran Liu$^{1,2}$, Yang Ye$^{1}$, Qianchao Zhu$^{1}$, Zane Cao$^{2}$, Yongchao He$^{2}$\thanks{Corresponding Author.} \\ 
$^1$Peking University, $^2$SCITIX (SGP) TECH PTE. LTD.\\
\texttt{liusr25@stu.pku.edu.cn} \\
}

\begin{document}
\maketitle

\begin{abstract}
Autoregressive decoding inherently limits the inference throughput of Large Language Model (LLM) due to its sequential dependency. \emph{Speculative decoding} mitigates this by verifying multiple predicted tokens in parallel, but its efficiency remains constrained by what we identify as \emph{verification heterogeneity}---the uneven difficulty of verifying different speculative candidates. In practice, a small subset of high-confidence predictions accounts for most successful verifications, yet existing methods treat all candidates uniformly, leading to redundant computation. 
We present \textbf{HeteroSpec}, a \textbf{hetero}geneity-adaptive \textbf{spec}ulative decoding framework that allocates verification effort in proportion to candidate uncertainty. HeteroSpec estimates verification complexity using a lightweight entropy-based quantifier, partitions candidates via a data-driven stratification policy, and dynamically tunes speculative depth and pruning thresholds through coordinated optimization. 
Across five benchmarks and four LLMs, HeteroSpec delivers an average \textbf{4.24$\times$} decoding speedup over state-of-the-art methods such as EAGLE-3, while preserving exact output distributions. Crucially, HeteroSpec requires no model retraining and remains compatible with other inference optimizations, making it a practical direction for improving speculative decoding efficiency.
\end{abstract}
\section{Introduction}

Autoregressive decoding serves as the foundation for modern large language models (LLMs), enabling high-quality text generation across diverse applications such as dialogue systems, summarization, and question answering~\cite{brown2020language, touvron2023llama, vaswani2023attentionneed, achiam2023gpt}. However, this sequential generation paradigm introduces a significant computational bottleneck, as each token requires a complete forward pass through the target model~\cite{kasai2021deepencodershallowdecoder, shazeer2019fasttransformerdecodingwritehead}. Developing inference acceleration techniques that preserve output quality and distributional correctness is therefore essential for scalable LLM deployment.

Speculative decoding has emerged as an effective approach to mitigate this sequential bottleneck~\cite{chen2023acceleratinglargelanguagemodel, leviathan2023fastinferencetransformersspeculative}. This paradigm employs a smaller \emph{draft model} to propose candidate token sequences, which are subsequently validated in parallel by the target model, reducing sequential forward passes while maintaining exact distributional guarantees through rejection sampling. Recent advances have explored diverse strategies: Medusa~\cite{cai2024medusasimplellminference} uses parallel prediction heads, EAGLE-2~\cite{li2024eagle2fasterinferencelanguage} introduces confidence-guided dynamic draft trees, and EAGLE-3~\cite{li2025eagle3scalinginferenceacceleration} leverages multi-layer feature aggregation with relaxed training constraints to improve draft quality.

Despite this progress, speculative decoding still faces a fundamental challenge stemming from the \textbf{dynamic and heterogeneous nature} of token prediction. 
As characterized by Zipf's Law~\cite{zipf1949human}, natural language exhibits a highly skewed distribution: a small set of high-frequency patterns constitutes the bulk of text, while a long tail of complex, low-frequency structures proves computationally challenging. Consequently, text generation is far from uniformly predictable; it constantly transitions between highly frequent, easily anticipatable patterns and complex, low-frequency structures, leading to dynamically fluctuating "decoding difficulty" during generation.

Current dynamic methods~\cite{brown2024dynamicdepthdecodingfaster,zhang2024draftmodelknowsstop,huang2024specdecboostingspeculativedecoding,zhang2024adaeagleoptimizingspeculativedecoding} recognize this need for adaptation, employing confidence metrics or trained predictors to control drafting length and stopping criteria. However, these approaches primarily focus on optimizing draft generation while the verification phase—which constitutes 67-90\% of total computation—receives limited attention for fine-grained adaptation. Moreover, static thresholds and pre-trained predictors struggle with the context complexity of language, making adaptive controls fragile across diverse generation scenarios.

To understand this challenge, we empirically profile the draft acceptance process in EAGLE-3 (\S\ref{sec:Observations}). Our analysis reveals a pronounced \emph{verification heterogeneity} in draft acceptance outcomes: a small fraction of high-confidence, top-ranked draft candidates are disproportionately responsible for accepted tokens and contribute significantly to acceleration, while the majority yield minimal or no accepted prefixes. This highlights the inefficiency of uniformly processing all candidates and strongly suggests that dynamically allocating computational resources, particularly the computationally expensive verification effort, based on predicted confidence and linguistic complexity, can substantially improve efficiency by prioritizing the most promising candidates.

Motivated by these empirical insights, we introduce \textbf{HeteroSpec} (\textbf{Hetero}geneity-Adaptive \textbf{Spec}ulative Decoding), a framework that addresses the verification bottleneck through complexity-aware adaptive optimization. HeteroSpec comprises three synergistic components that collectively transform uniform speculation into heterogeneity-adaptive resource allocation. First, it introduces a \emph{contextual complexity quantification} module that assesses real-time generation predictability using a novel cumulative meta-path Top-$K$ entropy metric. Second, based on this complexity score, an \emph{adaptive decision framework} employs data-driven entropy stratification to partition the generation process into distinct, actionable regimes. Finally, \emph{coordinated adaptive optimization} mechanisms leverage these regimes to dynamically adjust speculative depth, prune unpromising candidates, and select efficient computational graphs, thereby allocating verification resources where they are most effective.

Across five representative benchmarks spanning diverse task categories and four open-source LLMs,
HeteroSpec achieves a \textbf{4.24}$\times$ average speedup, consistently outperforming the state-of-the-art EAGLE-3 across all scenarios while maintaining exact distributional guarantees. This improvement costs less than $1\%$ additional overhead, yielding a net gain in overall inference speed. HeteroSpec is orthogonal to existing acceleration techniques and requires no model retraining for deployment. Our main contributions are:

\begin{itemize}
    \item \textbf{Empirical Characterization of Heterogeneity}: We demonstrate that a small fraction of high-confidence candidates disproportionately drive the majority of successful speculation. We establish this heterogeneity as a critical, previously overlooked performance bottleneck, thereby creating a new, data-driven foundation for optimization. (\S\ref{sec:Observations})
    \item \textbf{The HeteroSpec Framework}: We design and implement HeteroSpec, a novel adaptive decoding framework that integrates real-time complexity quantification, data-driven decision-making, and coordinated multi-level optimizations to dynamically allocate computational resources. (\S\ref{sec:Methodology})
    \item \textbf{Comprehensive Empirical Validation}: Experiments show that HeteroSpec significantly outperforms state-of-the-art methods across diverse models and benchmarks, establishing a new, more efficient standard for speculative decoding while providing in-depth analyses of its performance and robustness. (\S\ref{sec:Experiments})
\end{itemize}

\section{Preliminaries}

\subsection{Speculative Decoding}

Speculative decoding~\cite{chen2023acceleratinglargelanguagemodel, leviathan2023fastinferencetransformersspeculative} accelerates autoregressive LLM inference while preserving the exact target model probability distribution. A lightweight draft model proposes $k$ candidate tokens $\hat{T}_{j+1:j+k}$ following prefix $T_{1:j}$, which the target model processes in parallel. In the \emph{verification stage}, tokens are validated sequentially: $\hat{t}_{j+i}$ is accepted with probability $A_{j+i} = \min \left(1, \frac{p_{j+i}(\hat{t}_{j+i})}{\hat{p}_{j+i}(\hat{t}_{j+i})}\right)$, where $p$ and $\hat{p}$ denote target and draft distributions. Upon rejection, a token is sampled from residual distribution $p_{j+i} - \hat{p}_{j+i}$ to maintain fidelity, subsequent drafts are discarded, and decoding resumes from $j+i$.

\subsection{EAGLEs and its Draft Tree Construction}

Building on standard speculative decoding, the EAGLE family significantly advances LLM inference acceleration. Its key innovation is \textbf{dynamic draft tree construction}, introduced in EAGLE-2~\cite{li2024eagle2fasterinferencelanguage}. This approach moves beyond fixed-length drafts by adaptively proposing candidate sequences in tree structures through two stages: \textit{Expansion} and \textit{Reranking}. EAGLE-3~\cite{li2025eagle3scalinginferenceacceleration} refines this framework, enhancing predictive power by removing feature loss constraints and incorporating multi-layer information. The dynamic tree construction process in EAGLE-2 consists of the following two phases:

\begin{itemize}

\item \textbf{Expansion}: EAGLE-2 constructs preliminary draft tree $T_1$ by expanding nodes with Top-$k$ tokens from draft model distribution $\hat{p}$. Expansion prioritizes branches with high estimated global acceptance values $V_i$, defined as the product of acceptance probabilities along the path from root to node $i$. True acceptance probability $A_j$ is approximated by draft model confidence score $c_j$, yielding $V_i \approx \prod_{t_j \in \text{Path}(\text{root}, i)} c_j$. Expansion is limited by maximum depth $d$.

\item \textbf{Reranking}: All $T_1$ nodes are re-evaluated by global acceptance values $V_i$. Top-$N$ nodes with highest $V_i$ form pruned subtree $T_2$, maintaining validity since $V_i$ is bounded by ancestors. $T_2$ undergoes parallel target model verification using tree-mask attention to compute all token probabilities $p$ simultaneously.

\end{itemize}

\section{Observations}
\label{sec:Observations}

\begin{figure*}[ht]
  \centering
  \includegraphics[width=0.95\linewidth]{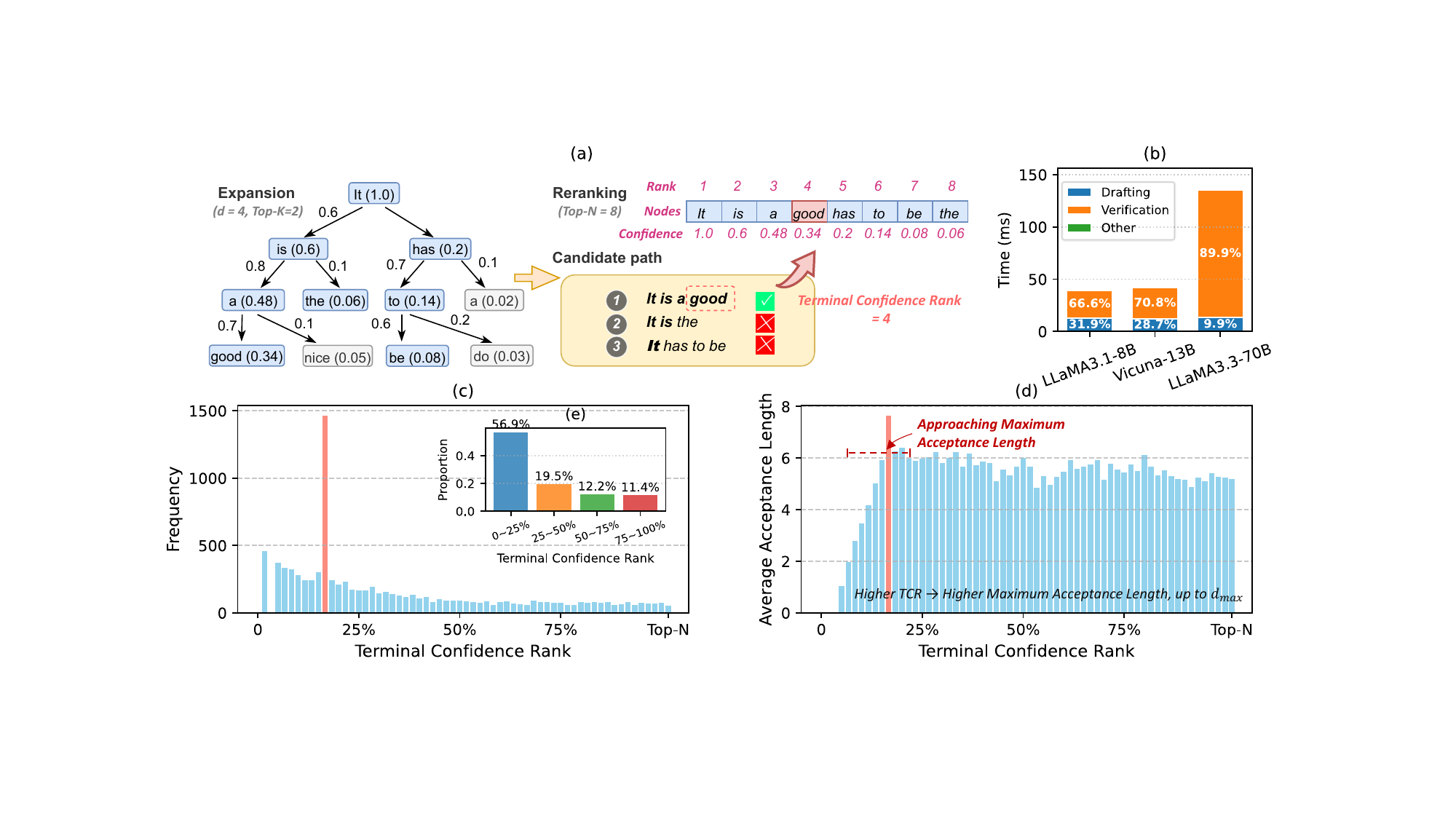}
 \caption{Key empirical observations with EAGLE-3. (a) Illustration of the Terminal Confidence Rank (TCR). (b) Breakdown of runtime overhead during single-turn speculative decoding for models of different sizes. (c) Distribution of Terminal Confidence Rank within the Top-$N$ draft candidates (with prominent values highlighted in orange). (d) Correlation between Average Acceptance Length and Terminal Confidence Rank. The initial rise reflects increasing maximum acceptance length as TCR grows, until reaching the maximum draft depth. (e) Quantile analysis of the Terminal Confidence Rank distribution, showing concentration within the top percentiles of Top-$N$.}
  \label{fig:motivation}
\end{figure*}
\subsection{The Verification Bottleneck}

The target model verification stage constitutes the primary computational bottleneck in speculative decoding.
As illustrated in Figure~\ref{fig:motivation}(b), this stage consumes a staggering 67-90\% of the total runtime across different model sizes. 
This substantial cost stems from the large size and complexity of target models. Consequently, improving overall inference efficiency critically depends on optimizing this stage, which can be achieved through two principles: (1) reducing the frequency of required verifications, and (2) decreasing the computational cost of each verification pass.

\subsection{Characterizing Heterogeneity}

To analyze successful draft acceptance dynamics, we introduce the metric of \textbf{Terminal Confidence Rank (TCR)}. As depicted in Figure~\ref{fig:motivation}(a), TCR is defined as the rank (among the Top-$N$ candidate sequences generated by the reranking phase) of the longest prefix ultimately accepted by the target model in a given decoding iteration.

Empirical analysis using EAGLE3-LLAMA3.1-8B on MT Bench~\cite{Zheng2023JudgingLW} reveals significant heterogeneity in the speculative decoding process (we provide additional experimental evidence of this heterogeneity across more models and datasets in Appendix~\ref{app:mtcrs}). Figure~\ref{fig:motivation}(c) and (e) show that Terminal Confidence Rank is heavily concentrated among the top 25\% of Top-$N$ candidates. Furthermore, Figure~\ref{fig:motivation}(d) shows a strong correlation between lower TCRs and longer average accepted lengths, often approaching the maximum acceptance length.
These findings indicate that sequences originating from high-confidence, top-ranked draft candidates are substantially more likely to be accepted and yield greater length gains.

This observed heterogeneity aligns with the nature of language and phenomena like Zipf's Law, whereby a small number of high-frequency patterns (such as common words and punctuation) make up the majority of natural language text. Simple, high-frequency linguistic patterns are more accurately predicted by the draft model, resulting in higher confidence (lower TCR) and higher acceptance probabilities leading to longer accepted prefixes. Conversely, complex or low-frequency structures are harder to predict, resulting in lower confidence (higher TCR) and shorter accepted lengths. As draft models improve, they expand the scope of patterns that can be reliably predicted, effectively reclassifying previously complex structures as simple, predictable patterns, thereby potentially amplifying this heterogeneity effect.

\subsection{Implications for Further Optimization}

Our empirical analysis reveals a pronounced heterogeneity in the success of draft candidates. We observe that \textbf{a small fraction of high-confidence paths generates the vast majority of successfully accepted tokens}, as shown in Figure~\ref{fig:motivation}(c-e). These high-potential paths are often characterized by a low TCR. Conversely, a large volume of draft candidates contributes little to no accepted sequence length, representing a significant source of redundant computation during the expensive verification stage. This clear asymmetry between a draft's potential and its final outcome is the central insight for further optimization.

Leveraging this insight, a clear optimization strategy emerges: \textbf{dynamically} allocate verification resources by prioritizing \textbf{high-potential draft paths}. Rather than treating all candidates uniformly, this approach concentrates the target model's computational budget on the branches most likely to yield long accepted sequences. This alignment of verification effort with the empirical likelihood of success directly maximizes the return on investment for each verification call. 
\section{Methodology}
\label{sec:Methodology}

\begin{figure*}[ht]
  \centering
  \includegraphics[width=0.97\linewidth]{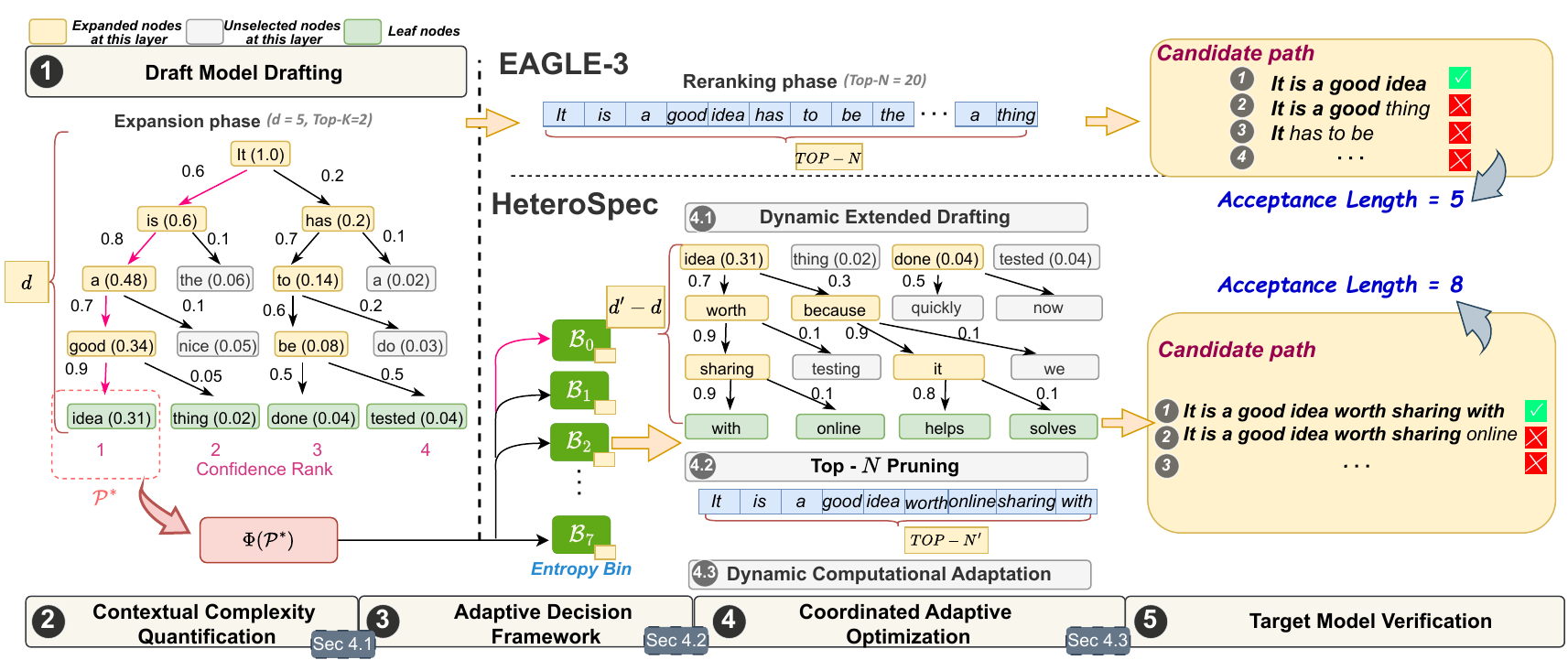}
  \caption{Illustration of the HeteroSpec framework, where \ding{173}, \ding{174}, and \ding{175} represent our three unique modules. We demonstrate the main differences between HeteroSpec and EAGLE-3 in the inference pipeline using an example of an EAGLE drafting tree with Top-$K$=2, Top-$N$=20, and Depth=5.
}
  \label{fig:method}
\end{figure*}

HeteroSpec comprises three synergistic components: Section~\ref{sec:41} introduces contextual complexity quantification for real-time predictability assessment; Section~\ref{sec:42} presents an adaptive decision framework that stratifies complexity patterns through data-driven partitioning; Section~\ref{sec:43} develops coordinated adaptive optimization mechanisms that dynamically allocate computational resources based on complexity assessment. Figure~\ref{fig:method} illustrates the unified framework.

\subsection{Contextual Complexity Quantification}
\label{sec:41}
The fundamental challenge in adaptive speculative decoding lies in distinguishing between high-predictability contexts that enable aggressive speculation and complex contexts requiring conservative approaches. 
To address this challenge, we introduce the \textbf{Cumulative Meta-Path Top-$K$ Entropy} as our complexity oracle. For a candidate speculation path $\mathcal{P} = (x_1, x_2, \ldots, x_T)$, we define:
{\fontsize{10pt}{14pt}\selectfont
\begin{equation}
\Phi(\mathcal{P}) = \sum_{t=1}^{T} \mathcal{I}_t(\mathcal{P}) = -\sum_{t=1}^{T} \sum_{i=1}^{\mathrm{Top}\text{-}K} \tilde{p}_{t,i} \log \tilde{p}_{t,i}
\end{equation}}
where $\mathcal{I}_t(\mathcal{P})$ captures the instantaneous uncertainty at position $t$, and $\tilde{p}_{t,i}$ represents the normalized probabilities of the Top-$K$ tokens. 

This design is motivated by two key principles: (1) Empirical foundation—simple, well-structured contexts exhibit highly skewed probability distributions with low per-step entropy, making cumulative entropy an effective discriminator for predictable generation patterns; (2) Computational efficiency—the Top-$K$ approximation ensures $\mathcal{O}(\mathrm{Top}\text{-}K \cdot T)$ complexity suitable for real-time assessment.

For the operational metric, we focus on the candidate path $\mathcal{P}^\ast$ with the highest final-token confidence, i.e., $\mathcal{P}^\ast = \arg\max_{\mathcal{P}} p^{(\lvert \mathcal{P} \rvert)}_{T, \mathrm{Top}\text{-}1}$. We compute $\Phi(\mathcal{P}^\ast)$ as the confidence indicator for the speculation tree at that step. We provide additional experimental validation of this metric's effectiveness in Appendix~\ref{app:cco}.

\subsection{Adaptive Decision Framework}
\label{sec:42}

The continuous nature of contextual complexity requires a principled method to discretize the complexity space into actionable decision regions that enable systematic resource allocation policies. Manual threshold specification lacks adaptability and fails to capture the nuanced relationships between complexity patterns and speculation outcomes.

We establish a \textbf{data-driven decision framework} that learns optimal complexity boundaries from empirical speculation trajectories, leveraging the interpretability and discretization strengths of Classification and Regression Tree (CART) algorithms~\cite{brei}. We train a 3-layer CART regression tree on a large corpus drawn from the ShareGPT dataset, which was originally utilized for draft model pre-training. From this corpus, we extract fully accepted draft paths to construct our training dataset $\mathcal{D} = \{(\mathbf{x}^{(i)}, y^{(i)})\}_{i=1}^{N}$, where $\mathbf{x}^{(i)} = \Phi(\mathcal{P}^{*(i)})$ represents the cumulative meta-path Top-$K$ entropy and $y^{(i)}$ denotes the Terminal Confidence Rank for the path $\mathcal{P}^*$. For each potential split threshold $s$, the framework minimizes within-partition variance of success outcomes:
{\fontsize{10pt}{14pt}\selectfont
\begin{equation}
\mathcal{L}(s) = \sum_{j \in \{\text{left}, \text{right}\}} \frac{|D_j|}{|D|} \sum_{(\mathbf{x}, y) \in D_j} (y - \mu_j)^2
\end{equation}}
where $D_j$ represents the resulting partitions and $\mu_j$ denotes the average rank in each partition.

Through recursive application of this optimization criterion, we construct a 3-layer decision tree yielding $8$ non-overlapping entropy bins $\{\mathcal{B}_0, \mathcal{B}_1, \ldots, \mathcal{B}_7\}$. We concentrate optimization efforts on the low-entropy bins $\mathcal{B}_l = \bigcup_{i=0}^{2} \mathcal{B}_i$, which exhibit the highest potential for aggressive speculation strategies and align with Zipf's law observations about high-frequency pattern dominance in natural language.

The shallow decision tree achieves \textbf{sub-3-second training times} while maintaining computational efficiency. Once trained, the decision tree is automatically integrated into the system and \textbf{generalizes to all task types without task-specific retraining or fine-tuning}, enabling efficient and fully automated deployment. This data-driven approach eliminates manual hyperparameter tuning while ensuring theoretical consistency through empirical risk minimization. Further analysis of decision tree selection and configuration sensitivity is provided in Appendices~\ref{app:cartdt} and~\ref{app:granularity}.

\subsection{Coordinated Adaptive Optimization}
\label{sec:43}

Having established complexity quantification and stratification, we implement three synergistic mechanisms that transform uniform speculation into adaptive resource allocation. Figure~\ref{fig:method} presents an inference pipeline, highlighting the main differences between EAGLE-3 and HeteroSpec.

\textbf{Dynamic Extended Drafting.}: For contexts assigned to low-entropy bins, we extend speculation depth based on complexity assessment:
\(d'(\mathcal{B}_i) = d_{\text{base}} + \Delta(\mathcal{B}_i)\)
where $\Delta(\mathcal{B}_i) = \alpha - i$ provides complexity-aware depth extension. Given the persistent prevalence of high-frequency and structurally simple patterns in natural language, when the current draft path falls into a low-entropy bin, subsequent speculative tokens are also likely to remain within low-entropy regions. In this scenario, increasing the speculative depth extends expected length of accepted segments, thereby reducing the total number of verifications required. Even in rare cases where the speculative extension departs from the low-entropy regime, the verification cost is only paid once for the entire segment, amortizing the computational overhead.

\textbf{Top-$N$ Pruning}: Rather than uniformly processing all candidates, we dynamically adjust the verification set size:
\(N'(\mathcal{B}_i) = \gamma_i \cdot N_{\text{default}} + \Delta(\mathcal{B}_i)\)
where $\gamma_i$ implements conservative upper bounds ($\gamma_1 = 0.3$, $\gamma_2 = 0.6$, $\gamma_3 = 1.0$) for the low-entropy bins. This mapping avoids the brittleness and complexity of manual hyperparameter tuning by adopting a conservative upper-bound policy: by retaining a relatively large quantile of candidates for each bin, we reliably capture the most likely accepted branches while significantly reducing unnecessary verification on unlikely candidates.

\textbf{Dynamic Computational Adaptation}: The varying speculation configurations introduce per-iteration variations in computation graph structure. To maintain inference efficiency under such dynamic control flow, we employ just-in-time compilation to generate and cache specialized computation graphs $G(\mathcal{B}_i)$ for distinct complexity regions, enabling effective operator fusion and computational reuse. This ensures that high-throughput inference is preserved, even as speculative depth and candidate set size dynamically change during decoding.

Quantitative overhead analysis confirms that our approach introduces negligible computational and memory costs (see Appendix~\ref{app:overhead}).
\section{Experiments}
\label{sec:Experiments}
\textbf{Tasks}. To demonstrate the generality of HeteroSpec, we conduct a comprehensive evaluation without task-specific fine-tuning across five representative benchmarks spanning key task categories: multi-turn dialogue (MT Bench~\cite{Zheng2023JudgingLW}), code generation (HumanEval~\cite{chen2021evaluating}), mathematical reasoning (GSM8K~\cite{cobbe2021training}), instruction following (Alpaca~\cite{alpaca}), and summarization (CNN/Daily Mail~\cite{nallapati2016abstractive}).

\textbf{Models}. We evaluate four representative LLMs spanning diverse model scales: Vicuna 13B~\cite{vicuna2023}, LLaMA-Instruct 3.1 8B, LLaMA-Instruct 3.3 70B~\cite{dubey2024llama}, and DeepSeek-R1-Distill-LLaMA 8B~\cite{deepseek2025deepseek}. Experiments on 8B/13B models utilize $ 1\times$ NVIDIA H20-3e 141G GPU, while the 70B model requires $ 2\times$ H20-3e GPUs due to memory constraints. Additionally, experimental results on other GPU types (A800 80G, L40 40G) are provided in the Appendix~\ref{app:moregpus}.

\textbf{Metrics}. 
HeteroSpec is intended to reduce the verification cost of the target model; therefore, we introduce two device-independent metrics: total validation calls and total verification tokens. In addition, HeteroSpec preserves the target model architecture and maintains strict acceptance conditions, generation quality evaluation is unnecessary.
\begin{itemize}
    \item \textbf{Speedup Ratio}: Actual acceleration compared to vanilla autoregressive decoding.
    \item \textbf{Average Acceptance Length ($\tau$)}: Mean tokens generated per drafting-verification cycle.
    \item \textbf{Total Validation Calls (Calls) / Total Verification Tokens(Tokens)}: The former counts target model invocations during decoding, while the latter quantifies cumulative tokens processed across all validation steps. Together, these metrics capture the computational overhead of candidate verification.
\end{itemize}

\textbf{Baseline}. We benchmark against state-of-the-art EAGLE-3~\cite{li2025eagle3scalinginferenceacceleration}, adopting identical hyperparameters (Depth, Top-$K$, Top-$N$) from the official implementation. Since this study does not involve training the draft model, and to avoid the confounding effect of randomness on the interpretation of our method's effectiveness, only the case of temperature=0 is considered by default throughout the following analysis. Detailed analysis of the temperature=1 case is provided in Appendix~\ref{app:t1}.

\subsection{Effectiveness}
\begin{table*}[ht]
  \centering
\resizebox{\linewidth}{!}
    {\small
    \begin{tabular}{cccccccccccccc}
    \toprule
          &       & \multicolumn{2}{c}{MT-bench} & \multicolumn{2}{c}{HumanEval} & \multicolumn{2}{c}{GSM8K} & \multicolumn{2}{c}{Alpaca} & \multicolumn{2}{c}{CNN/DM} & \multicolumn{2}{c}{Mean} \\
    \midrule
    Model & Method & Speedup & $\tau$     & Speedup & $\tau$     & Speedup & $\tau$     & Speedup & $\tau$     & Speedup & $\tau$     & Speedup & $\tau$ \\
      &  & Calls & Tokens     & Calls & Tokens    & Calls & Tokens     & Calls & Tokens     & Calls & Tokens     & Calls & Tokens \\
    \midrule

\multirow{4}{*}{V 13B}
      & \multirow{2}{*}{EAGLE-3} & 3.22$\times$ & 6.56  & 3.78$\times$ & 7.73  & 3.31$\times$ & 6.40  & 3.31$\times$ & 6.51  & 2.87$\times$ & 6.54  & 3.30$\times$ & 6.75 \\
      & &  6257 &  327026 &  3427 &  171598 &  2421 &  123480 &  2058 &  108192 &  1975 &  101185 &  3228 &  166296 \\
\cmidrule(lr){2-14}
      & \multirow{2}{*}{HeteroSpec} & \textbf{3.53$\times$} & \textbf{7.10}  & \textbf{4.45$\times$} & \textbf{9.05}  & \textbf{3.42$\times$} & \textbf{6.67}  & \textbf{3.45$\times$} & \textbf{6.70}  & \textbf{3.04$\times$} & \textbf{7.07}  & \textbf{3.58$\times$} & \textbf{7.32} \\
      & &  \textbf{5937} &  \textbf{278942} &  \textbf{2948} &  \textbf{119753} &  \textbf{2321} &  \textbf{109919} &  \textbf{2011} &  \textbf{101778} &  \textbf{1860} &  \textbf{86147} &  \textbf{3015} &  \textbf{139308} \\
    \midrule

\multirow{4}{*}{L31 8B}
      & \multirow{2}{*}{EAGLE-3} & 3.43$\times$ & 6.15  & 3.95$\times$ & 6.83  & 3.69$\times$ & 6.32  & 3.79$\times$ & 6.84  & 2.92$\times$ & 5.39  & 3.56$\times$ & 6.31 \\
      & &  8687 &  541384 &  4118 &  248272 &  2465 &  150568 &  2563 &  161660 &  4009 &  245204 &  4368 &  269418 \\
\cmidrule(lr){2-14}
      & \multirow{2}{*}{HeteroSpec} & \textbf{3.74$\times$} & \textbf{6.44}  & \textbf{4.43$\times$} & \textbf{7.46}  & \textbf{3.81$\times$} & \textbf{6.53}  & \textbf{3.99$\times$} & \textbf{7.08}  & \textbf{3.11$\times$} & \textbf{5.47}  & \textbf{3.82$\times$} & \textbf{6.60} \\
      & &  \textbf{8357} &  \textbf{489985} &  \textbf{3799} &  \textbf{196029} &  \textbf{2379} &  \textbf{141487} &  \textbf{2462} &  \textbf{151801} &  \textbf{3878} &  \textbf{232705} &  \textbf{4175} &  \textbf{242401} \\
    \midrule

\multirow{4}{*}{L33 70B}
      & \multirow{2}{*}{EAGLE-3} & 5.33$\times$ & 5.64  & 6.49$\times$ & 6.67  & 5.95$\times$ & 6.22  & 5.89$\times$ & 6.58  & 4.36$\times$ & 5.01  & 5.60$\times$ & 6.02 \\
      & &  10099 &  499187 &  4702 &  225788 &  2372 &  115996 &  2890 &  146264 &  4345 &  211970 &  4882 &  239841 \\
\cmidrule(lr){2-14}
      & \multirow{2}{*}{HeteroSpec} & \textbf{5.38$\times$} & \textbf{5.80}  & \textbf{6.76$\times$} & \textbf{7.02}  & \textbf{6.06$\times$} & \textbf{6.40}  & \textbf{6.02$\times$} & \textbf{6.82}  & \textbf{4.43$\times$} & \textbf{5.04}  & \textbf{5.73$\times$} & \textbf{6.22} \\
      & &  \textbf{9696} &  \textbf{444761} &  \textbf{4478} &  \textbf{169249} &  \textbf{2313} &  \textbf{103342} &  \textbf{2829} &  \textbf{134361} &  \textbf{4340} &  \textbf{204626} &  \textbf{4731} &  \textbf{211268} \\
    \midrule
    
\multirow{4}{*}{DSL 8B}
      & \multirow{2}{*}{EAGLE-3} & 3.50$\times$ & 5.88  & 3.87$\times$ & 6.58  & 4.12$\times$ & 7.09  & 3.25$\times$ & 5.62  & 2.82$\times$ & 5.02  & 3.51$\times$ & 6.04 \\
      & &  6965 &  425803 &  6331 &  387689 &  4809 &  295413 &  7086 &  434004 &  8139 &  499494 &  6666 &  408481 \\
\cmidrule(lr){2-14}
      & \multirow{2}{*}{HeteroSpec} & \textbf{3.78$\times$} & \textbf{6.12}  & \textbf{4.19$\times$} & \textbf{6.82}  & \textbf{4.64$\times$} & \textbf{7.89}  & \textbf{3.56$\times$} & \textbf{5.75}  & \textbf{2.99$\times$} & \textbf{5.07}  & \textbf{3.83$\times$} & \textbf{6.33} \\
      & &  \textbf{6787} &  \textbf{398080} &  \textbf{6135} &  \textbf{356517} &  \textbf{4374} &  \textbf{226139} &  \textbf{6962} &  \textbf{418119} &  \textbf{8085} &  \textbf{487250} &  \textbf{6469} &  \textbf{377221} \\
    \bottomrule
    \end{tabular}
      }
    \caption{Speedup ratios, average acceptance lengths ($\tau$), total validation calls (Calls), and total verification tokens (Tokens) of different methods. V represents Vicuna, L31 represents LLaMA-Instruct 3.1, L33 represents LLaMA-Instruct 3.3, and DSL represents DeepSeek-R1-Distill-LLaMA. }
  \label{tab:experiments}
\end{table*}

Table~\ref{tab:experiments} demonstrates that HeteroSpec consistently outperforms EAGLE-3 across all evaluated datasets and LLMs on four key metrics: speedup ratio, average acceptance length, total validation calls, and total verification tokens. Our method achieves an average 4.24× speedup over autoregressive decoding with an average acceptance length of 6.62.

In the code generation task (HumanEval), HeteroSpec demonstrates the most significant performance gains. This superiority arises from the high predictability of predominantly templated code structures, with EAGLE-3 typically exhibiting high confidence and being assigned to low-entropy bins. Through dynamic extended drafting, HeteroSpec facilitates extended token sequence acceptance, achieving a 5.65\% reduction in target model verifications. Additionally, Top-$N$ pruning yields a 22.79\% decrease in verification tokens while preserving generation accuracy.

For the summarization task (CNN/DM), increased output diversity and unpredictability diminish draft-target model alignment, leading to shorter average accepted lengths for EAGLE-3, with more frequent assignments to high-entropy bins. Consequently, HeteroSpec's performance gains in these tasks are comparatively modest, and its behavior partially converges towards EAGLE-3. Nevertheless, these results demonstrate HeteroSpec's robust adaptability and effectiveness across diverse models and tasks, consistently reducing target model validation overhead.

HeteroSpec maintains orthogonality to existing draft model optimization techniques, achieving correspondingly greater performance gains as draft models improve without any additional cost. 
EAGLE-3 further enhances mathematical reasoning capability by additionally training the DeepSeekR1-Distill-LLaMA 8B model on the OpenThoughts114k-math dataset, resulting in even more pronounced improvements for HeteroSpec on GSM8K—demonstrating this scaling effect.

\subsection{Ablation Study}

\begin{table*}[ht]
  \centering
\resizebox{\linewidth}{!}
  {\footnotesize
    \begin{tabular}{c|cccc|cccc|cccc}
      \toprule
      & \multicolumn{4}{c|}{LLaMA3.1-8B}
      & \multicolumn{4}{c|}{Vicuna-13B}
      & \multicolumn{4}{c}{LLaMA3.3-70B} \\
      \cmidrule{2-13}
      Method
      & Speedup & $\tau$ & Calls & Tokens
      & Speedup & $\tau$ & Calls & Tokens
      & Speedup & $\tau$ & Calls & Tokens \\
      \midrule
      EAGLE-3   & 3.95$\times$ & 6.83 & 4118 & 248272 & 3.78$\times$ & 7.73 & 3427 & 171598 & 6.49$\times$ & 6.67 & 4702 & 225788 \\
      +DED       & 4.35$\times$ & 7.54 & 3757 & 226796 & 4.38$\times$ & 9.13 & 2900 & 144893 & 6.71$\times$ & 7.18 & 4367 & 209620 \\
      +TNP       & 4.31$\times$ & 7.46 & 3799 & 196029 & 4.36$\times$ & 9.05 & 2948 & 119753 & 6.68$\times$ & 7.02 & 4428 & 169249 \\
      +DCA     & 4.43$\times$ & 7.46 & 3799 & 196029 & 4.45$\times$ & 9.05 & 2948 & 119753 & 6.76$\times$ & 7.02 & 4428 & 169249 \\
      \bottomrule
    \end{tabular}
  }
    \caption{
  Ablation study on HumanEval dataset across three models of different sizes. DED (Dynamic Extended Drafting), TNP (Top-$N$ Pruning), and DCA (Dynamic Computational Adaptation) are incrementally integrated; each row represents the ablation result after adding the corresponding module to the previous configuration. 
  }
  \label{tab:ablation}
\end{table*}

To systematically assess each optimization, we conducted ablation experiments progressively integrating our three adaptive mechanisms upon EAGLE-3. Table~\ref{tab:ablation} demonstrates that Dynamic Extended Drafting yields the most substantial performance gains. Its underlying mechanism is that, when cumulative meta-path Top-$K$ entropy is low, the probability that an initially drafted token is accepted at all layers increases significantly. This enables subsequent tokens' verification costs to be amortized together with the initial draft tokens, regardless of whether these subsequent tokens fall into low-entropy bins. Given the prevalence of simple patterns in natural language, subsequent tokens typically remain within low-entropy regions, further amplifying this cost-sharing benefit. This mechanism increases average accepted length and reduces total verifications by 11.09\%, substantially alleviating the verification bottleneck.

Top-$N$ Pruning achieves a 19.85\% reduction in verification tokens with minimal impact on accepted length (0.11 decrease). This demonstrates that the strategy can select, at key decision points, the most likely-to-be-accepted critical paths dynamically and at minimal cost. Notably, although there is a slight 0.59\% decrease in the speedup metric, this reflects GPU underutilization in single-batch H20-3e scenarios rather than algorithmic limitations; In the discussion of complex multi-batch scenarios (see Section~\ref{sec:54}), this strategy reveals greater potential for acceleration. Dynamic Computational Adaptation mitigates performance degradation from computational graph variations introduced by the preceding optimizations. These three mechanisms collectively demonstrate the synergistic effectiveness of our contextual heterogeneity exploitation framework.

\subsection{Hyperparameter Study}

\begin{table*}[ht]
  \centering
\resizebox{\linewidth}{!}
  {\footnotesize
    \begin{tabular}{c|cccc|cccc|cccc}
      \toprule
      & \multicolumn{4}{c|}{LLaMA3.1-8B}
      & \multicolumn{4}{c|}{Vicuna-13B}
      & \multicolumn{4}{c}{LLaMA3.3-70B} \\
      \cmidrule{2-13}
      $\alpha$
      & Speedup & $\tau$ & Calls & Tokens
      & Speedup & $\tau$ & Calls & Tokens
      & Speedup & $\tau$ & Calls & Tokens \\
      \midrule
        EAGLE-3   & 3.95$\times$ & 6.83 & 4118 & 248272 & 3.78$\times$ & 7.73 & 3427 & 171598 & 6.49$\times$ & 6.67 & 4702 & 225788 \\
        $\left \lceil depth/2 \right \rceil - 1$          & 4.36$\times$ & 7.37 & 3846 & 198486 & 4.35$\times$ & 8.71 & 3049 & 126859 & 6.68$\times$ & 7.00 & 4482 & 169351 \\
      $\left \lceil depth/2 \right \rceil $   & 4.43$\times$ & 7.46 & 3799 & 196029 & 4.45$\times$ & 9.05 & 2948 & 119753 & 6.76$\times$ & 7.02 & 4478 & 169249 \\
      $\left \lceil depth/2 \right \rceil + 1$          & 4.38$\times$ & 7.51 & 3778 & 195053 & 4.44$\times$ & 9.23 & 2906 & 118657 & 6.71$\times$ & 7.01 & 4473 & 169151 \\
      \bottomrule
    \end{tabular}
  }
    \caption{Hyperparameter study ($\alpha$) on HumanEval dataset across three models of different sizes.}
  \label{tab:hyper}
\end{table*}

To investigate the hyperparameter $\alpha$ in Dynamic Extended Drafting, we conduct experiments with varying $\alpha$ values (see Table~\ref{tab:hyper}). We find that the speedup is maximized when $\alpha$ is set to $\left \lceil depth/2 \right \rceil$, while both increasing or decreasing $\alpha$ leads to reduced acceleration. In essence, this hyperparameter balances the additional drafting overhead against the reduction in verification cost. If $\alpha$ is too low, the draft model underutilizes its capacity in low-entropy regions, missing opportunities to accept longer token sequences. Conversely, if $\alpha$ is too high, the additional drafting overhead outweighs verification savings, leading to degraded performance. These results indicate that optimal $\alpha$ scales with draft model capability—as models improve, a larger $\alpha$ can yield greater speedups.

\section{Related Work}

Speculative decoding~\cite{chen2023acceleratinglargelanguagemodel, leviathan2023fastinferencetransformersspeculative} adopts a "drafting with lightweight model, verification with original model" paradigm for lossless acceleration. SpecInfer~\cite{Miao_2024} introduced tree-based attention for efficient parallel verification. REST~\cite{fu2024breaksequentialdependencyllm} and LLMA~\cite{yang2023inferencereferencelosslessacceleration} employ retrieval-based drafting. GLIDE~\cite{du2024glidecapelowhasslemethod} and MoA~\cite{zimmer2025mixtureattentionsspeculativedecoding} reuse target model KV cache. Medusa~\cite{cai2024medusasimplellminference}, Hydra~\cite{ankner2024hydrasequentiallydependentdraftheads}, EAGLE~\cite{li2025eaglespeculativesamplingrequires}, and EAGLE-3~\cite{li2025eagle3scalinginferenceacceleration} reuse target model feature representations. Other approaches~\cite{zhang2023draft,yi2024generation,elhoushi2024layer,liu2024kangaroo,sun2024triforce,svirschevski2024specexec} reuse partial target model weights. These approaches focus on training stronger draft models, which is orthogonal to our method and enables greater gains as draft performance improves.

Another research line explores dynamic draft structures. EAGLE-2~\cite{li2024eagle2fasterinferencelanguage} uses dynamic drafting trees with joint probability confidence. BiLD~\cite{kim2023speculativedecodingbiglittle}, Kangaroo~\cite{liu2024kangaroo}, DDD~\cite{brown2024dynamicdepthdecodingfaster}, and SVIP~\cite{zhang2024draftmodelknowsstop} introduce stopping metrics. SpecDec++~\cite{huang2024specdecboostingspeculativedecoding} and AdaEAGLE~\cite{zhang2024adaeagleoptimizingspeculativedecoding} train modules for early stopping or draft length prediction. C2T~\cite{huo2025c2tclassifierbasedtreeconstruction} proposes classifier-based tree to correct joint probability bias. However, these methods inadequately exploit linguistic heterogeneity for verification optimization, limiting performance, applicability, and timeliness. Our method adaptively optimizes computational resource allocation, achieving superior performance over existing approaches.
\section{Conclusion}
Based on the challenges posed by the heterogeneous statistical properties of natural language, this work presents HeteroSpec, a complexity-aware speculative decoding framework that addresses the verification bottleneck through adaptive resource allocation. Through cumulative meta-path Top-$K$ entropy quantification and data-driven stratification, HeteroSpec transforms uniform verification into adaptive optimization that dynamically allocates resources to high-confidence candidates. Extensive evaluation demonstrates that HeteroSpec achieves greater speedups than the state-of-the-art EAGLE-3, with negligible computational overhead, establishing the value of leveraging linguistic heterogeneity for efficient LLM inference. Future work is discussed in Appendix~\ref{app:Future}.
\section{Limitations}

A limitation of HeteroSpec is its current implementation within the state-of-the-art EAGLE framework, necessitating validation for generalizability to other speculative decoding methods. However, the core principle of dynamic resource allocation guided by linguistic heterogeneity is fundamentally orthogonal and broadly applicable.

\bibliography{custom}

\begin{thebibliography}{46}
\providecommand{\natexlab}[1]{#1}

\bibitem[{Achiam et~al.(2023)Achiam, Adler, Agarwal, Ahmad, Akkaya, Aleman, Almeida, Altenschmidt, Altman, Anadkat et~al.}]{achiam2023gpt}
Josh Achiam, Steven Adler, Sandhini Agarwal, Lama Ahmad, Ilge Akkaya, Florencia~Leoni Aleman, Diogo Almeida, Janko Altenschmidt, Sam Altman, Shyamal Anadkat, and 1 others. 2023.
\newblock Gpt-4 technical report.
\newblock \emph{arXiv preprint arXiv:2303.08774}.

\bibitem[{Ankner et~al.(2024)Ankner, Parthasarathy, Nrusimha, Rinard, Ragan-Kelley, and Brandon}]{ankner2024hydrasequentiallydependentdraftheads}
Zachary Ankner, Rishab Parthasarathy, Aniruddha Nrusimha, Christopher Rinard, Jonathan Ragan-Kelley, and William Brandon. 2024.
\newblock \href {https://arxiv.org/abs/2402.05109} {Hydra: Sequentially-dependent draft heads for medusa decoding}.
\newblock \emph{Preprint}, arXiv:2402.05109.

\bibitem[{Breiman et~al.(1984)Breiman, Friedman, Olshen, and Stone}]{brei}
L.~Breiman, Jerome~H. Friedman, Richard~A. Olshen, and C.~J. Stone. 1984.
\newblock Classification and regression trees.
\newblock In \emph{Proceedings of the Conference}.

\bibitem[{Brown et~al.(2024)Brown, Wang, Do, Mathew, and Yu}]{brown2024dynamicdepthdecodingfaster}
Oscar Brown, Zhengjie Wang, Andrea Do, Nikhil Mathew, and Cheng Yu. 2024.
\newblock \href {https://arxiv.org/abs/2409.00142} {Dynamic depth decoding: Faster speculative decoding for llms}.
\newblock \emph{Preprint}, arXiv:2409.00142.

\bibitem[{Brown et~al.(2020)Brown, Mann, Ryder, Subbiah, Kaplan, Dhariwal, Neelakantan, Shyam, Sastry, Askell, Agarwal, Herbert-Voss, Krueger, Henighan, Child, Ramesh, Ziegler, Wu, Winter, Hesse, Chen, Sigler, Litwin, Gray, Chess, Clark, Berner, McCandlish, Radford, Sutskever, and Amodei}]{brown2020language}
Tom~B. Brown, Benjamin Mann, Nick Ryder, Melanie Subbiah, Jared Kaplan, Prafulla Dhariwal, Arvind Neelakantan, Pranav Shyam, Girish Sastry, Amanda Askell, Sandhini Agarwal, Ariel Herbert-Voss, Gretchen Krueger, Tom Henighan, Rewon Child, Aditya Ramesh, Daniel~M. Ziegler, Jeffrey Wu, Clemens Winter, and 12 others. 2020.
\newblock Language models are few-shot learners.
\newblock In \emph{Proceedings of the 34th International Conference on Neural Information Processing Systems}, NIPS '20, Red Hook, NY, USA. Curran Associates Inc.

\bibitem[{Brysbaert(2019)}]{BRYSBAERT2019104047}
Marc Brysbaert. 2019.
\newblock \href {https://doi.org/10.1016/j.jml.2019.104047} {How many words do we read per minute? a review and meta-analysis of reading rate}.
\newblock \emph{Journal of Memory and Language}, 109:104047.

\bibitem[{Cai et~al.(2024)Cai, Li, Geng, Peng, Lee, Chen, and Dao}]{cai2024medusasimplellminference}
Tianle Cai, Yuhong Li, Zhengyang Geng, Hongwu Peng, Jason~D. Lee, Deming Chen, and Tri Dao. 2024.
\newblock \href {https://arxiv.org/abs/2401.10774} {Medusa: Simple llm inference acceleration framework with multiple decoding heads}.
\newblock \emph{Preprint}, arXiv:2401.10774.

\bibitem[{Chen et~al.(2023)Chen, Borgeaud, Irving, Lespiau, Sifre, and Jumper}]{chen2023acceleratinglargelanguagemodel}
Charlie Chen, Sebastian Borgeaud, Geoffrey Irving, Jean-Baptiste Lespiau, Laurent Sifre, and John Jumper. 2023.
\newblock \href {https://arxiv.org/abs/2302.01318} {Accelerating large language model decoding with speculative sampling}.
\newblock \emph{Preprint}, arXiv:2302.01318.

\bibitem[{Chen et~al.(2021)Chen, Tworek, Jun, Yuan, Pinto, Kaplan, Edwards, Burda, Joseph, Brockman et~al.}]{chen2021evaluating}
Mark Chen, Jerry Tworek, Heewoo Jun, Qiming Yuan, Henrique Ponde de~Oliveira Pinto, Jared Kaplan, Harri Edwards, Yuri Burda, Nicholas Joseph, Greg Brockman, and 1 others. 2021.
\newblock Evaluating large language models trained on code.
\newblock \emph{arXiv preprint arXiv:2107.03374}.

\bibitem[{Chiang et~al.(2023)Chiang, Li, Lin, Sheng, Wu, Zhang, Zheng, Zhuang, Zhuang, Gonzalez, Stoica, and Xing}]{vicuna2023}
Wei-Lin Chiang, Zhuohan Li, Zi~Lin, Ying Sheng, Zhanghao Wu, Hao Zhang, Lianmin Zheng, Siyuan Zhuang, Yonghao Zhuang, Joseph~E. Gonzalez, Ion Stoica, and Eric~P. Xing. 2023.
\newblock \href {https://lmsys.org/blog/2023-03-30-vicuna/} {Vicuna: An open-source chatbot impressing gpt-4 with 90\%* chatgpt quality}.

\bibitem[{Cobbe et~al.(2021)Cobbe, Kosaraju, Bavarian, Chen, Jun, Kaiser, Plappert, Tworek, Hilton, Nakano et~al.}]{cobbe2021training}
Karl Cobbe, Vineet Kosaraju, Mohammad Bavarian, Mark Chen, Heewoo Jun, Lukasz Kaiser, Matthias Plappert, Jerry Tworek, Jacob Hilton, Reiichiro Nakano, and 1 others. 2021.
\newblock Training verifiers to solve math word problems.
\newblock \emph{arXiv preprint arXiv:2110.14168}.

\bibitem[{DeepSeek-AI et~al.(2025)DeepSeek-AI, Yang, Zhang, Song, Zhang, Xu, Zhu, Ma, Wang, Bi et~al.}]{deepseek2025deepseek}
Daya~Guo DeepSeek-AI, Dejian Yang, Haowei Zhang, Junxiao Song, Ruoyu Zhang, Runxin Xu, Qihao Zhu, Shirong Ma, Peiyi Wang, Xiao Bi, and 1 others. 2025.
\newblock Deepseek-r1: Incentivizing reasoning capability in llms via reinforcement learning.
\newblock \emph{arXiv preprint arXiv:2501.12948}.

\bibitem[{Du et~al.(2024)Du, Jiang, Yuanchen, Wu, Yu, Li, Li, Xu, Nie, Tu, and You}]{du2024glidecapelowhasslemethod}
Cunxiao Du, Jing Jiang, Xu~Yuanchen, Jiawei Wu, Sicheng Yu, Yongqi Li, Shenggui Li, Kai Xu, Liqiang Nie, Zhaopeng Tu, and Yang You. 2024.
\newblock \href {https://arxiv.org/abs/2402.02082} {Glide with a cape: A low-hassle method to accelerate speculative decoding}.
\newblock \emph{Preprint}, arXiv:2402.02082.

\bibitem[{Dubey et~al.(2024)Dubey, Jauhri, Pandey, Kadian, Al-Dahle, Letman, Mathur, Schelten, Yang, Fan et~al.}]{dubey2024llama}
Abhimanyu Dubey, Abhinav Jauhri, Abhinav Pandey, Abhishek Kadian, Ahmad Al-Dahle, Aiesha Letman, Akhil Mathur, Alan Schelten, Amy Yang, Angela Fan, and 1 others. 2024.
\newblock The {Llama} 3 herd of models.
\newblock \emph{arXiv preprint arXiv:2407.21783}.

\bibitem[{Elhoushi et~al.(2024)Elhoushi, Shrivastava, Liskovich, Hosmer, Wasti, Lai, Mahmoud, Acun, Agarwal, Roman et~al.}]{elhoushi2024layer}
Mostafa Elhoushi, Akshat Shrivastava, Diana Liskovich, Basil Hosmer, Bram Wasti, Liangzhen Lai, Anas Mahmoud, Bilge Acun, Saurabh Agarwal, Ahmed Roman, and 1 others. 2024.
\newblock Layer skip: Enabling early exit inference and self-speculative decoding.
\newblock \emph{arXiv preprint arXiv:2404.16710}.

\bibitem[{Fu et~al.(2024)Fu, Bailis, Stoica, and Zhang}]{fu2024breaksequentialdependencyllm}
Yichao Fu, Peter Bailis, Ion Stoica, and Hao Zhang. 2024.
\newblock \href {https://arxiv.org/abs/2402.02057} {Break the sequential dependency of llm inference using lookahead decoding}.
\newblock \emph{Preprint}, arXiv:2402.02057.

\bibitem[{Huang et~al.(2024)Huang, Guo, and Wang}]{huang2024specdecboostingspeculativedecoding}
Kaixuan Huang, Xudong Guo, and Mengdi Wang. 2024.
\newblock \href {https://arxiv.org/abs/2405.19715} {Specdec++: Boosting speculative decoding via adaptive candidate lengths}.
\newblock \emph{Preprint}, arXiv:2405.19715.

\bibitem[{Huo et~al.(2025)Huo, Tan, Zhang, Cai, and Sun}]{huo2025c2tclassifierbasedtreeconstruction}
Feiye Huo, Jianchao Tan, Kefeng Zhang, Xunliang Cai, and Shengli Sun. 2025.
\newblock \href {https://arxiv.org/abs/2502.13652} {C2t: A classifier-based tree construction method in speculative decoding}.
\newblock \emph{Preprint}, arXiv:2502.13652.

\bibitem[{Kasai et~al.(2021)Kasai, Pappas, Peng, Cross, and Smith}]{kasai2021deepencodershallowdecoder}
Jungo Kasai, Nikolaos Pappas, Hao Peng, James Cross, and Noah~A. Smith. 2021.
\newblock \href {https://arxiv.org/abs/2006.10369} {Deep encoder, shallow decoder: Reevaluating non-autoregressive machine translation}.
\newblock \emph{Preprint}, arXiv:2006.10369.

\bibitem[{Kim et~al.(2023)Kim, Mangalam, Moon, Malik, Mahoney, Gholami, and Keutzer}]{kim2023speculativedecodingbiglittle}
Sehoon Kim, Karttikeya Mangalam, Suhong Moon, Jitendra Malik, Michael~W. Mahoney, Amir Gholami, and Kurt Keutzer. 2023.
\newblock \href {https://arxiv.org/abs/2302.07863} {Speculative decoding with big little decoder}.
\newblock \emph{Preprint}, arXiv:2302.07863.

\bibitem[{Leviathan et~al.(2023)Leviathan, Kalman, and Matias}]{leviathan2023fastinferencetransformersspeculative}
Yaniv Leviathan, Matan Kalman, and Yossi Matias. 2023.
\newblock \href {https://arxiv.org/abs/2211.17192} {Fast inference from transformers via speculative decoding}.
\newblock \emph{Preprint}, arXiv:2211.17192.

\bibitem[{Li et~al.(2024)Li, Wei, Zhang, and Zhang}]{li2024eagle2fasterinferencelanguage}
Yuhui Li, Fangyun Wei, Chao Zhang, and Hongyang Zhang. 2024.
\newblock \href {https://arxiv.org/abs/2406.16858} {Eagle-2: Faster inference of language models with dynamic draft trees}.
\newblock \emph{Preprint}, arXiv:2406.16858.

\bibitem[{Li et~al.(2025{\natexlab{a}})Li, Wei, Zhang, and Zhang}]{li2025eagle3scalinginferenceacceleration}
Yuhui Li, Fangyun Wei, Chao Zhang, and Hongyang Zhang. 2025{\natexlab{a}}.
\newblock \href {https://arxiv.org/abs/2503.01840} {Eagle-3: Scaling up inference acceleration of large language models via training-time test}.
\newblock \emph{Preprint}, arXiv:2503.01840.

\bibitem[{Li et~al.(2025{\natexlab{b}})Li, Wei, Zhang, and Zhang}]{li2025eaglespeculativesamplingrequires}
Yuhui Li, Fangyun Wei, Chao Zhang, and Hongyang Zhang. 2025{\natexlab{b}}.
\newblock \href {https://arxiv.org/abs/2401.15077} {Eagle: Speculative sampling requires rethinking feature uncertainty}.
\newblock \emph{Preprint}, arXiv:2401.15077.

\bibitem[{Li et~al.(2025{\natexlab{c}})Li, Chen, Delacourt, Oliaro, Wang, Chen, Lin, Yang, Zhang, Chen, Lai, Miao, and Jia}]{li2025adaserveslocustomizedllmserving}
Zikun Li, Zhuofu Chen, Remi Delacourt, Gabriele Oliaro, Zeyu Wang, Qinghan Chen, Shuhuai Lin, April Yang, Zhihao Zhang, Zhuoming Chen, Sean Lai, Xupeng Miao, and Zhihao Jia. 2025{\natexlab{c}}.
\newblock \href {https://arxiv.org/abs/2501.12162} {Adaserve: Slo-customized llm serving with fine-grained speculative decoding}.
\newblock \emph{Preprint}, arXiv:2501.12162.

\bibitem[{Lin et~al.(2024)Lin, Han, Zhang, Yang, Yang, Chen, and Qiu}]{lin2024parrotefficientservingllmbased}
Chaofan Lin, Zhenhua Han, Chengruidong Zhang, Yuqing Yang, Fan Yang, Chen Chen, and Lili Qiu. 2024.
\newblock \href {https://arxiv.org/abs/2405.19888} {Parrot: Efficient serving of llm-based applications with semantic variable}.
\newblock \emph{Preprint}, arXiv:2405.19888.

\bibitem[{Liu et~al.(2024{\natexlab{a}})Liu, Tang, Liu, Ni, Han, and Wang}]{liu2024kangaroo}
Fangcheng Liu, Yehui Tang, Zhenhua Liu, Yunsheng Ni, Kai Han, and Yunhe Wang. 2024{\natexlab{a}}.
\newblock Kangaroo: Lossless self-speculative decoding via double early exiting.
\newblock \emph{arXiv preprint arXiv:2404.18911}.

\bibitem[{Liu et~al.(2024{\natexlab{b}})Liu, Daniel, Hu, Kwon, Li, Mo, Cheung, Deng, Stoica, and Zhang}]{liu2024optimizingspeculativedecodingserving}
Xiaoxuan Liu, Cade Daniel, Langxiang Hu, Woosuk Kwon, Zhuohan Li, Xiangxi Mo, Alvin Cheung, Zhijie Deng, Ion Stoica, and Hao Zhang. 2024{\natexlab{b}}.
\newblock \href {https://arxiv.org/abs/2406.14066} {Optimizing speculative decoding for serving large language models using goodput}.
\newblock \emph{Preprint}, arXiv:2406.14066.

\bibitem[{Miao et~al.(2024)Miao, Oliaro, Zhang, Cheng, Wang, Zhang, Wong, Zhu, Yang, Shi, Shi, Chen, Arfeen, Abhyankar, and Jia}]{Miao_2024}
Xupeng Miao, Gabriele Oliaro, Zhihao Zhang, Xinhao Cheng, Zeyu Wang, Zhengxin Zhang, Rae Ying~Yee Wong, Alan Zhu, Lijie Yang, Xiaoxiang Shi, Chunan Shi, Zhuoming Chen, Daiyaan Arfeen, Reyna Abhyankar, and Zhihao Jia. 2024.
\newblock \href {https://doi.org/10.1145/3620666.3651335} {Specinfer: Accelerating large language model serving with tree-based speculative inference and verification}.
\newblock In \emph{Proceedings of the 29th ACM International Conference on Architectural Support for Programming Languages and Operating Systems, Volume 3}, ASPLOS ’24, page 932–949. ACM.

\bibitem[{Nallapati et~al.(2016)Nallapati, Zhou, Gulcehre, Xiang et~al.}]{nallapati2016abstractive}
Ramesh Nallapati, Bowen Zhou, Caglar Gulcehre, Bing Xiang, and 1 others. 2016.
\newblock Abstractive text summarization using sequence-to-sequence rnns and beyond.
\newblock \emph{arXiv preprint arXiv:1602.06023}.

\bibitem[{Sadhukhan et~al.(2025)Sadhukhan, Chen, Chen, Tiwari, Lai, Shi, Yen, May, Chen, and Chen}]{sadhukhan2025magicdecbreakinglatencythroughputtradeoff}
Ranajoy Sadhukhan, Jian Chen, Zhuoming Chen, Vashisth Tiwari, Ruihang Lai, Jinyuan Shi, Ian En-Hsu Yen, Avner May, Tianqi Chen, and Beidi Chen. 2025.
\newblock \href {https://arxiv.org/abs/2408.11049} {Magicdec: Breaking the latency-throughput tradeoff for long context generation with speculative decoding}.
\newblock \emph{Preprint}, arXiv:2408.11049.

\bibitem[{Shazeer(2019)}]{shazeer2019fasttransformerdecodingwritehead}
Noam Shazeer. 2019.
\newblock \href {https://arxiv.org/abs/1911.02150} {Fast transformer decoding: One write-head is all you need}.
\newblock \emph{Preprint}, arXiv:1911.02150.

\bibitem[{Sun et~al.(2024)Sun, Chen, Yang, Tian, and Chen}]{sun2024triforce}
Hanshi Sun, Zhuoming Chen, Xinyu Yang, Yuandong Tian, and Beidi Chen. 2024.
\newblock Triforce: Lossless acceleration of long sequence generation with hierarchical speculative decoding.
\newblock \emph{arXiv preprint arXiv:2404.11912}.

\bibitem[{Svirschevski et~al.(2024)Svirschevski, May, Chen, Chen, Jia, and Ryabinin}]{svirschevski2024specexec}
Ruslan Svirschevski, Avner May, Zhuoming Chen, Beidi Chen, Zhihao Jia, and Max Ryabinin. 2024.
\newblock Specexec: Massively parallel speculative decoding for interactive llm inference on consumer devices.
\newblock \emph{arXiv preprint arXiv:2406.02532}.

\bibitem[{Taori et~al.(2023)Taori, Gulrajani, Zhang, Dubois, Li, Guestrin, Liang, and Hashimoto}]{alpaca}
Rohan Taori, Ishaan Gulrajani, Tianyi Zhang, Yann Dubois, Xuechen Li, Carlos Guestrin, Percy Liang, and Tatsunori~B Hashimoto. 2023.
\newblock Alpaca: A strong, replicable instruction-following model.
\newblock \emph{Stanford Center for Research on Foundation Models. https://crfm. stanford. edu/2023/03/13/alpaca. html}, 3(6):7.

\bibitem[{Touvron et~al.(2023)Touvron, Lavril, Izacard, Martinet, Lachaux, Lacroix, Rozière, Goyal, Hambro, Azhar, Rodriguez, Joulin, Grave, and Lample}]{touvron2023llama}
Hugo Touvron, Thibaut Lavril, Gautier Izacard, Xavier Martinet, Marie-Anne Lachaux, Timothée Lacroix, Baptiste Rozière, Naman Goyal, Eric Hambro, Faisal Azhar, Aurelien Rodriguez, Armand Joulin, Edouard Grave, and Guillaume Lample. 2023.
\newblock \href {https://arxiv.org/abs/2302.13971} {Llama: Open and efficient foundation language models}.
\newblock \emph{Preprint}, arXiv:2302.13971.

\bibitem[{Vaswani et~al.(2023)Vaswani, Shazeer, Parmar, Uszkoreit, Jones, Gomez, Kaiser, and Polosukhin}]{vaswani2023attentionneed}
Ashish Vaswani, Noam Shazeer, Niki Parmar, Jakob Uszkoreit, Llion Jones, Aidan~N. Gomez, Lukasz Kaiser, and Illia Polosukhin. 2023.
\newblock \href {https://arxiv.org/abs/1706.03762} {Attention is all you need}.
\newblock \emph{Preprint}, arXiv:1706.03762.

\bibitem[{Yang et~al.(2023)Yang, Ge, Wang, Jiao, Jiang, Yang, Majumder, and Wei}]{yang2023inferencereferencelosslessacceleration}
Nan Yang, Tao Ge, Liang Wang, Binxing Jiao, Daxin Jiang, Linjun Yang, Rangan Majumder, and Furu Wei. 2023.
\newblock \href {https://arxiv.org/abs/2304.04487} {Inference with reference: Lossless acceleration of large language models}.
\newblock \emph{Preprint}, arXiv:2304.04487.

\bibitem[{Yi et~al.(2024)Yi, Lin, Li, Ning, Yu, and Xiao}]{yi2024generation}
Hanling Yi, Feng Lin, Hongbin Li, Peiyang Ning, Xiaotian Yu, and Rong Xiao. 2024.
\newblock Generation meets verification: Accelerating large language model inference with smart parallel auto-correct decoding.
\newblock \emph{arXiv preprint arXiv:2402.11809}.

\bibitem[{Zhang et~al.(2023)Zhang, Wang, Li, Shou, Chen, Chen, and Mehrotra}]{zhang2023draft}
Jun Zhang, Jue Wang, Huan Li, Lidan Shou, Ke~Chen, Gang Chen, and Sharad Mehrotra. 2023.
\newblock Draft \& verify: Lossless large language model acceleration via self-speculative decoding.
\newblock \emph{arXiv preprint arXiv:2309.08168}.

\bibitem[{Zhang et~al.(2024{\natexlab{a}})Zhang, Wang, Ma, Zhu, Chen, Lan, and Yu}]{zhang2024adaeagleoptimizingspeculativedecoding}
Situo Zhang, Hankun Wang, Da~Ma, Zichen Zhu, Lu~Chen, Kunyao Lan, and Kai Yu. 2024{\natexlab{a}}.
\newblock \href {https://arxiv.org/abs/2412.18910} {Adaeagle: Optimizing speculative decoding via explicit modeling of adaptive draft structures}.
\newblock \emph{Preprint}, arXiv:2412.18910.

\bibitem[{Zhang et~al.(2024{\natexlab{b}})Zhang, Xu, Liang, Chen, He, Wang, and Tu}]{zhang2024draftmodelknowsstop}
Ziyin Zhang, Jiahao Xu, Tian Liang, Xingyu Chen, Zhiwei He, Rui Wang, and Zhaopeng Tu. 2024{\natexlab{b}}.
\newblock \href {https://arxiv.org/abs/2411.18462} {Draft model knows when to stop: A self-verification length policy for speculative decoding}.
\newblock \emph{Preprint}, arXiv:2411.18462.

\bibitem[{Zheng et~al.(2023)Zheng, Chiang, Sheng, Zhuang, Wu, Zhuang, Lin, Li, Li, Xing, Zhang, Gonzalez, and Stoica}]{Zheng2023JudgingLW}
Lianmin Zheng, Wei-Lin Chiang, Ying Sheng, Siyuan Zhuang, Zhanghao Wu, Yonghao Zhuang, Zi~Lin, Zhuohan Li, Dacheng Li, Eric~P. Xing, Haotong Zhang, Joseph~E. Gonzalez, and Ion Stoica. 2023.
\newblock \href {https://api.semanticscholar.org/CorpusID:259129398} {Judging llm-as-a-judge with mt-bench and chatbot arena}.
\newblock \emph{ArXiv}, abs/2306.05685.

\bibitem[{Zhong et~al.(2024)Zhong, Liu, Chen, Hu, Zhu, Liu, Jin, and Zhang}]{zhong2024distservedisaggregatingprefilldecoding}
Yinmin Zhong, Shengyu Liu, Junda Chen, Jianbo Hu, Yibo Zhu, Xuanzhe Liu, Xin Jin, and Hao Zhang. 2024.
\newblock \href {https://arxiv.org/abs/2401.09670} {Distserve: Disaggregating prefill and decoding for goodput-optimized large language model serving}.
\newblock \emph{Preprint}, arXiv:2401.09670.

\bibitem[{Zimmer et~al.(2025)Zimmer, Gritta, Lampouras, Ammar, and Wang}]{zimmer2025mixtureattentionsspeculativedecoding}
Matthieu Zimmer, Milan Gritta, Gerasimos Lampouras, Haitham~Bou Ammar, and Jun Wang. 2025.
\newblock \href {https://arxiv.org/abs/2410.03804} {Mixture of attentions for speculative decoding}.
\newblock \emph{Preprint}, arXiv:2410.03804.

\bibitem[{Zipf(1949)}]{zipf1949human}
George~Kingsley Zipf. 1949.
\newblock \emph{Human Behavior and the Principle of Least Effort}.
\newblock Addison-Wesley, Reading, MA.

\end{thebibliography}
\clearpage
\appendix
\section{Appendix}
\subsection{Additional Experiments on Heterogeneity}
\label{app:mtcrs}

\begin{figure*}[!htbp]
  \centering
  \includegraphics[width=1.00\linewidth]{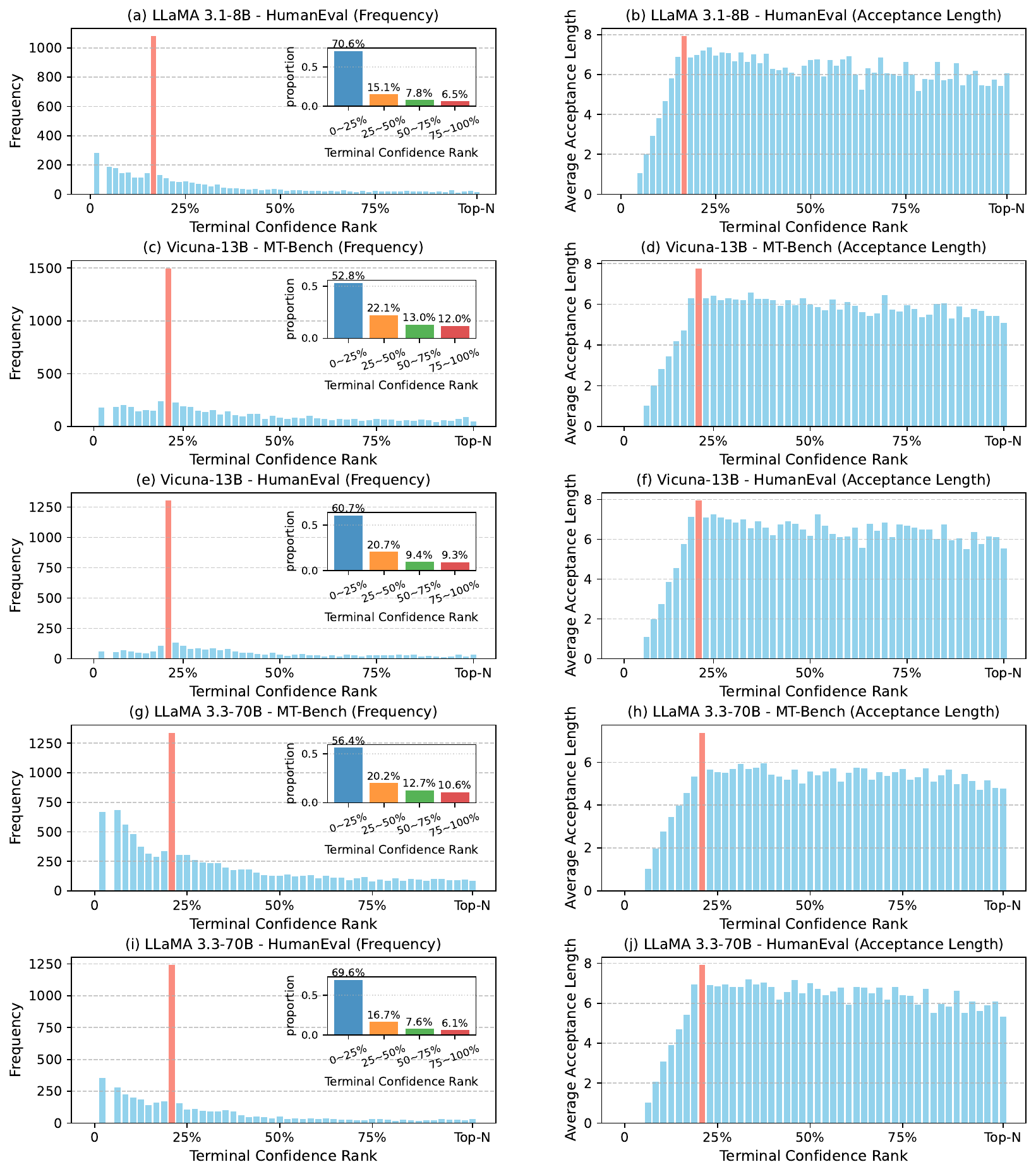}
 \caption{Terminal Confidence Rank analysis across multiple models and datasets. Left panels: Distribution of Terminal Confidence Rank within Top-$N$ draft candidates (bar chart, with prominent values highlighted in orange), with inset quantile analysis showing concentration within top percentiles. Right panels: Correlation between Average Acceptance Length and Terminal Confidence Rank (bar chart). (a,b) LLaMA 3.1-8B on HumanEval; (c,d) Vicuna-13B on MT-Bench; (e,f) Vicuna-13B on HumanEval; (g,h) LLaMA 3.3-70B on MT-Bench; (i,j) LLaMA 3.3-70B on HumanEval.}
  \label{fig:hete}
\end{figure*}

To further validate the heterogeneity observations presented in Section 3, we conduct comprehensive experiments across multiple models and datasets. Figure~\ref{fig:hete} presents Terminal Confidence Rank (TCR) analysis results for five additional model-dataset combinations: LLaMA-Instruct 3.1-8B on HumanEval, Vicuna-13B and LLaMA-Instruct 3.3-70B on both MT Bench and HumanEval.

The left panels show that Terminal Confidence Rank is heavily concentrated among the top 25\% of Top-$N$ candidates across all model-dataset combinations. Furthermore, the right panels show a strong correlation between lower TCRs and longer average accepted lengths, often approaching the maximum draft depth. These findings demonstrate remarkable consistency in heterogeneity patterns across different model scales (from 8B to 70B parameters) and task types, confirming that a small fraction of high-confidence candidates drives the majority of successful speculation. This indicates that heterogeneity in draft acceptance is an intrinsic property of speculative decoding rather than an artifact of specific model configurations.

Interestingly, the heterogeneity effect appears more pronounced in code generation tasks (HumanEval) compared to dialogue tasks (MT Bench). This aligns with the structured and predictable nature of code syntax patterns, which enables draft models to achieve higher confidence and longer acceptance sequences for well-formed code structures. These comprehensive results establish that the heterogeneity phenomenon identified in our main analysis is not limited to specific model-dataset combinations but represents a fundamental characteristic of speculative decoding across diverse settings. This broad validation strengthens the motivation for our heterogeneity-adaptive optimization approach and confirms that the observed patterns provide a robust foundation for the HeteroSpec framework.

\subsection{Contextual Complexity Oracle Validation}
\label{app:cco}
\begin{figure*}[!htbp]
  \centering
  \includegraphics[width=1.00\linewidth]{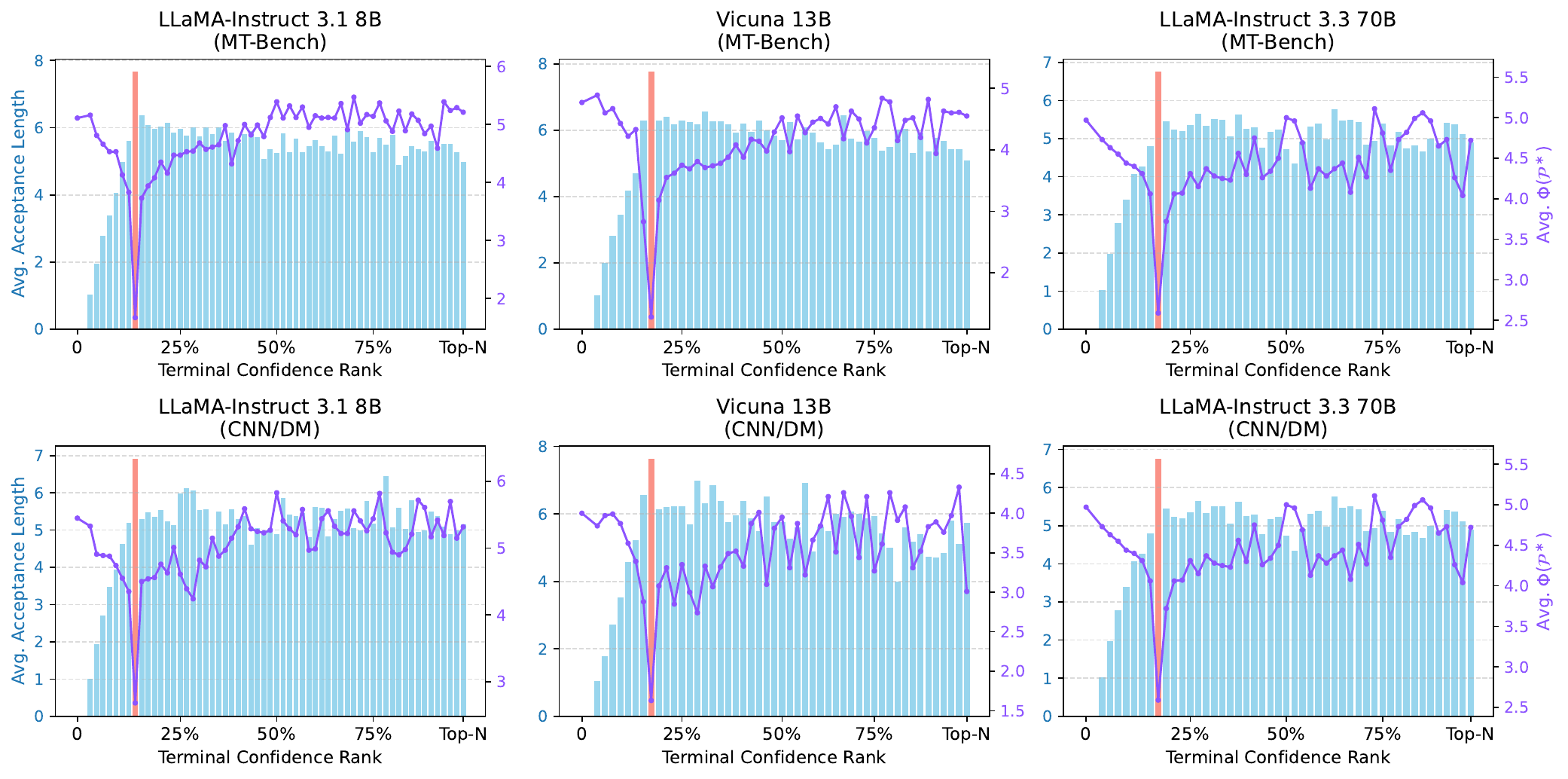}
 \caption{Validation results of the proposed Cumulative Meta-Path Top-$K$ Entropy metric across LLaMA 8B, Vicuna 13B, and LLaMA 70B models on MT-Bench and CNN/Daily Mail datasets. For each subplot, the x-axis shows Terminal Confidence Rank (TCR), the left y-axis indicates average acceptance length (bar chart, with prominent values highlighted in orange), and the right y-axis shows average Cumulative Meta-Path Top-$K$ Entropy (line plot).}
  \label{fig:cco}
\end{figure*}

To empirically validate the effectiveness of the proposed Cumulative Meta-Path Top-$K$ Entropy as a contextual complexity oracle, we conduct extensive experiments on three models: LLaMA-Instruct 3.1 8B, Vicuna 13B, and LLaMA-Instruct 3.3 70B using two representative datasets, MT Bench and CNN/Daily Mail. Results are shown in Figure~\ref{fig:cco}, which plots average acceptance length and average Cumulative Meta-Path Top-$K$ Entropy with respect to Terminal Confidence Rank (TCR).

We find that low-entropy regions, corresponding to simple and predictable contexts identified by our metric, consistently yield substantially longer average acceptance lengths. This demonstrates that the metric effectively isolates high-confidence contexts amenable to aggressive speculation. In contrast, higher entropy is associated with much greater variability in acceptance length, highlighting the unpredictability of complex contexts.

Notably, as illustrated in Figure~\ref{fig:motivation}(a), high-confidence, low-complexity contexts make up the majority of cases, mirroring the dominance of simple linguistic patterns in natural language. By providing a reliable and actionable measure of contextual complexity, our metric reveals significant optimization potential within these frequent low-entropy regions. Accordingly, our adaptive optimization framework is specifically designed to prioritize and exploit these regions for maximal inference acceleration.

\subsection{Rationale for Adopting CART Decision Trees for Complexity Stratification}
\label{app:cartdt}
To systematically partition the contextual complexity space, we adopt a data-driven CART decision tree framework in lieu of neural networks or other continuous models. This choice is theoretically motivated by both the structural properties uncovered by our contextual complexity oracle and the operational requirements of efficient speculative decoding.

Comprehensive experimental analysis of the cumulative meta-path Top-$K$ entropy metric (see Appendix~\ref{app:cco}) reveals a distinct asymmetry in the complexity landscape: low-entropy regions correspond to stable, highly predictable contexts with consistently high draft acceptance, while high-entropy regions lack discernible structure and exhibit wide variability.  Such step-like, non-smooth transitions in system behavior are well-suited to models with strong capacity for learning discrete partition boundaries. Decision trees, particularly the CART algorithm, are able to capture these sharp entropy thresholds in a principled, data-driven manner, directly optimizing for empirical risk minimization on partitioned complexity outcomes. In contrast, neural networks are fundamentally biased toward smooth function approximation, which hinders their ability to sharply distinguish abrupt changes or encode actionable thresholding policies.

From a systems perspective, most state-of-the-art draft models prioritize minimal inference latency and computational cost—typically employing a single-layer transformer for maximal efficiency. Introducing an additional neural network for online complexity partitioning would incur significant overhead, contradicting the primary goal of fast speculative decoding. In comparison, evaluating a shallow decision tree imposes only a few conditional checks per draft step, resulting in negligible runtime cost. As evidenced by our quantitative analysis in Appendix~\ref{app:overhead}, this approach introduces virtually no latency overhead in practice. Moreover, decision tree training itself is highly efficient; constructing a shallow CART tree typically completes within seconds.
\subsection{Entropy Stratification Granularity Analysis}
\label{app:granularity}

To validate the robustness of our data-driven entropy stratification module, we investigate the consistency of low-entropy region identification across different stratification granularities. We employ the \textbf{Jaccard similarity coefficient} as our primary evaluation metric, defined as:

\begin{equation}
J(A, B) = \frac{|A \cap B|}{|A \cup B|}
\end{equation}
where $A$ and $B$ represent the sets of samples identified as low-entropy by different stratification depths. Using our 3-layer stratification as the reference baseline, we examine consistency with deeper architectures by maintaining proportional low-entropy region selection across all granularities. Specifically, we compute $J(\mathcal{S}_{3L}, \mathcal{S}_{dL})$ for $d \in \{4, 5, 6\}$, where $\mathcal{S}_{dL}$ denotes the low-entropy sample set identified by $d$-layer stratification with equivalent proportional selection criteria. This coefficient ranges from 0 (no overlap) to 1 (perfect agreement), providing a quantitative measure of consistency across granularity variations.

\begin{table}[h]
\centering
\resizebox{\linewidth}{!}
{\footnotesize
\begin{tabular}{cccc}
\toprule
Model & 3L-4L Overlap & 3L-5L Overlap & 3L-6L Overlap \\
\midrule
L31 8B & 1.00 & 0.89 & 0.98 \\
V 13B & 0.96 & 0.97 & 1.00 \\
L33 70B & 1.00 & 0.98 & 0.96 \\
\bottomrule
\end{tabular}}
\caption{Low-entropy region overlap consistency across stratification granularities}
\label{tab:entropy-overlap}
\end{table}

As shown in Table~\ref{tab:entropy-overlap}, our analysis reveals remarkably high overlap consistency across all model configurations, with Jaccard coefficients consistently exceeding 0.89. Notably, multiple configurations achieve perfect agreement (coefficient = 1.00), demonstrating that different stratification granularities often converge to identical low-entropy region identification. This consistency indicates that critical complexity boundaries are intrinsic to model behavioral patterns rather than stratification artifacts, with deeper architectures providing minimal additional discriminative information for low-entropy detection.

These findings establish the granularity insensitivity of our entropy stratification framework, demonstrating three key properties: (1) Structural Stability: our method captures fundamental complexity patterns that persist across granularity variations; (2) Computational Efficiency: deeper stratification provides negligible gains while increasing overhead, demonstrating that our 3-layer design achieves an effective trade-off between accuracy and computational cost; (3) Universal Applicability: consistent patterns across model scales confirm broad applicability of our approach. This empirical evidence strongly justifies our architectural choice and establishes the robustness of complexity-aware speculative decoding.


\subsection{Runtime Overhead Profiling}
\label{app:overhead}

\begin{table}[ht]
\centering
\setlength{\tabcolsep}{1mm}
{\footnotesize
\begin{tabular}{lcccccc}
\toprule
\multicolumn{7}{c}{MT Bench} \\
\midrule
Model & Method & Draft & Verify & Add. & Total & Mem \\
\midrule
\multirow{2}{*}{L31 8B} & EAGLE-3 & 98.1 & 207.5 & - & 305.6 & 18.9 \\
& HeteroSpec & 103.1 & 194.2 & 2.3 & 299.6 & 18.9 \\
\cmidrule{1-7}
\multirow{2}{*}{V 13B} & EAGLE-3 & 79.2 & 207.9 & - & 287.1 & 29.8 \\
& HeteroSpec & 83.0 & 190.5 & 1.8 & 275.3 & 29.8 \\
\cmidrule{1-7}
\multirow{2}{*}{L33 70B} & EAGLE-3 & 133.3 & 1080.7 & - & 1214.0 & 143.8 \\
& HeteroSpec & 140.3 & 1030.6 & 3.1 & 1174.0 & 143.8 \\
\midrule
\multicolumn{7}{c}{HumanEval} \\
\midrule
Model & Method & Draft & Verify & Add. & Total & Mem \\
\midrule
\multirow{2}{*}{L31 8B} & EAGLE-3 & 45.3 & 93.7 & - & 139.0 & 18.8 \\
& HeteroSpec & 49.5 & 84.7 & 1.2 & 135.4 & 18.8 \\
\cmidrule{1-7}
\multirow{2}{*}{V 13B} & EAGLE-3 & 41.7 & 108.7 & - & 150.4 & 29.8 \\
& HeteroSpec & 43.7 & 91.4 & 0.9 & 136.0 & 29.8 \\
\cmidrule{1-7}
\multirow{2}{*}{L33 70B} & EAGLE-3 & 60.0 & 477.8 & - & 537.8 & 142.6 \\
& HeteroSpec & 68.9 & 434.2 & 1.3 & 504.4 & 142.6 \\
\bottomrule
\end{tabular}
}
\caption{Runtime overhead and memory profiling across benchmarks and models. Time values are reported in seconds (s) as cumulative totals across all test cases in each dataset. Memory usage is measured in GB and represents average runtime memory. Draft, Verify, and Add. represent drafting time, verification time, and additional overhead introduced by HeteroSpec's complexity assessment and adaptive optimization, respectively. V represents Vicuna, L31 represents LLaMA-Instruct 3.1, L33 represents LLaMA-Instruct 3.3.}
\label{tab:overhead}
\end{table}

To quantitatively assess the computational cost of our proposed approach, we conduct comprehensive runtime profiling experiments across MT Bench and HumanEval datasets. Our analysis decomposes total inference time into three distinct components: drafting overhead, verification overhead, and additional computational cost introduced by HeteroSpec's complexity assessment and adaptive optimization mechanisms. We evaluate this decomposition across different model scales to capture performance characteristics comprehensively.

The results in Table~\ref{tab:overhead} demonstrate that HeteroSpec introduces minimal additional overhead, ranging from 0.88 to 3.1 seconds across all configurations. This represents only 0.26\% to 0.87\% of total inference time, confirming the efficiency of our lightweight complexity quantification framework. Runtime memory usage remains essentially identical across both methods, demonstrating that our approach does not introduce significant memory overhead. While drafting overhead increases modestly due to dynamic depth extension, this is substantially offset by significant verification overhead reductions of 4.6\% to 15.9\%. Since verification constitutes 67.4\% to 89.0\% of total runtime—the primary computational bottleneck in speculative decoding—these reductions drive meaningful system-level gains.

Our profiling validates that HeteroSpec achieves performance improvements through targeted optimization with negligible additional overhead, confirming the efficiency of our data-driven complexity assessment and adaptive resource allocation, establishing that contextual heterogeneity can be effectively leveraged for speculative decoding acceleration.

\subsection{More results on different GPUs}
\label{app:moregpus}

\begin{table*}[!ht ]
  \centering
\resizebox{\linewidth}{!}
  {\footnotesize
    \begin{tabular}{cccccccccccccc}
    \toprule
          &       & \multicolumn{2}{c}{MT-bench} & \multicolumn{2}{c}{HumanEval} & \multicolumn{2}{c}{GSM8K} & \multicolumn{2}{c}{Alpaca} & \multicolumn{2}{c}{CNN/DM} & \multicolumn{2}{c}{Mean} \\
    \midrule
    Model & Method & Speedup & $\tau$     & Speedup & $\tau$     & Speedup & $\tau$     & Speedup & $\tau$     & Speedup & $\tau$     & Speedup & $\tau$ \\
      &  & Calls & Tokens     & Calls & Tokens    & Calls & Tokens     & Calls & Tokens     & Calls & Tokens     & Calls & Tokens \\
    \midrule

\multirow{4}{*}{V 13B}
      & \multirow{2}{*}{EAGLE-3} & 4.01$\times$ & 6.55  & 4.78$\times$ & 7.74  & 3.95$\times$ & 6.40  & 3.99$\times$ & 6.50  & 3.60$\times$ & 6.55  & 4.07$\times$ & 6.75 \\
      & &  6267 &  327516 &  3424 &  171451 &  2433 &  124068 &  2064 &  108486 &  1983 &  101577 &  3234 &  166620 \\
\cmidrule(lr){2-14}
      & \multirow{2}{*}{HeteroSpec} & \textbf{4.32$\times$} & \textbf{7.09}  & \textbf{5.41$\times$} & \textbf{9.04}  & \textbf{4.15$\times$} & \textbf{6.67}  & \textbf{4.14$\times$} & \textbf{6.70}  & \textbf{3.81$\times$} & \textbf{7.05}  & \textbf{4.37$\times$} & \textbf{7.31} \\
      & &  \textbf{5925} &  \textbf{281774} &  \textbf{2971} &  \textbf{119797} &  \textbf{2322} &  \textbf{110137} &  \textbf{2013} &  \textbf{101910} &  \textbf{1840} &  \textbf{84932} &  \textbf{3014} &  \textbf{139710} \\
    \midrule

\multirow{4}{*}{L31 8B}
      & \multirow{2}{*}{EAGLE-3} & 3.83$\times$ & 6.16  & 4.47$\times$ & 6.83  & 3.92$\times$ & 6.34  & 4.19$\times$ & 6.83  & 3.18$\times$ & 5.39  & 3.92$\times$ & 6.31 \\
      & &  8704 &  542387 &  4124 &  248626 &  2477 &  151276 &  2573 &  162604 &  3973 &  243080 &  4370 &  269595 \\
\cmidrule(lr){2-14}
      & \multirow{2}{*}{HeteroSpec} & \textbf{4.04$\times$} & \textbf{6.47}  & \textbf{4.75$\times$} & \textbf{7.36}  & \textbf{4.11$\times$} & \textbf{6.53}  & \textbf{4.39$\times$} & \textbf{7.04}  & \textbf{3.37$\times$} & \textbf{5.45}  & \textbf{4.13$\times$} & \textbf{6.57} \\
      & &  \textbf{8319} &  \textbf{485731} &  \textbf{3845} &  \textbf{198391} &  \textbf{2380} &  \textbf{141658} &  \textbf{2502} &  \textbf{153617} &  \textbf{3897} &  \textbf{233771} &  \textbf{4189} &  \textbf{242634} \\
    \midrule

\multirow{4}{*}{L33 70B}
      & \multirow{2}{*}{EAGLE-3} & 4.24$\times$ & 5.66  & 5.10$\times$ & 6.69  & 4.70$\times$ & 6.23  & 4.80$\times$ & 6.58  & 3.61$\times$ & 5.01  & 4.49$\times$ & 6.03 \\
      & &  10017 &  494910 &  4643 &  223015 &  2367 &  115761 &  2892 &  146358 &  4392 &  214179 &  4862 &  238845 \\
\cmidrule(lr){2-14}
      & \multirow{2}{*}{HeteroSpec} & \textbf{4.35$\times$} & \textbf{5.80}  & \textbf{5.39$\times$} & \textbf{7.04}  & \textbf{4.81$\times$} & \textbf{6.39}  & \textbf{4.88$\times$} & \textbf{6.76}  & \textbf{3.66$\times$} & \textbf{5.03}  & \textbf{4.62$\times$} & \textbf{6.20} \\
      & &  \textbf{9747} &  \textbf{447511} &  \textbf{4449} &  \textbf{168115} &  \textbf{2320} &  \textbf{103464} &  \textbf{2838} &  \textbf{134851} &  \textbf{4362} &  \textbf{205676} &  \textbf{4743} &  \textbf{211923} \\
    \midrule
    
\multirow{4}{*}{DSL 8B}
      & \multirow{2}{*}{EAGLE-3} & 3.67$\times$ & 5.85  & 4.12$\times$ & 6.57  & 4.67$\times$ & 7.09  & 3.47$\times$ & 5.64  & 3.09$\times$ & 5.02  & 3.80$\times$ & 6.03 \\
      & &  7013 &  428458 &  6353 &  388987 &  4814 &  295708 &  7084 &  433001 &  8118 &  498255 &  6676 &  408882 \\
\cmidrule(lr){2-14}
      & \multirow{2}{*}{HeteroSpec} & \textbf{3.88$\times$} & \textbf{6.12}  & \textbf{4.41$\times$} & \textbf{6.82}  & \textbf{5.08$\times$} & \textbf{7.89}  & \textbf{3.71$\times$} & \textbf{5.75}  & \textbf{3.25$\times$} & \textbf{5.06}  & \textbf{4.07$\times$} & \textbf{6.33} \\
      & &  \textbf{6797} &  \textbf{398972} &  \textbf{6132} &  \textbf{356129} &  \textbf{4381} &  \textbf{228799} &  \textbf{6924} &  \textbf{415689} &  \textbf{8079} &  \textbf{485283} &  \textbf{6463} &  \textbf{376974} \\
    \bottomrule
    \end{tabular}
    }
      \caption{Speedup ratios, average acceptance lengths ($\tau$), total validation calls (Calls), and total verification tokens (Tokens) of different methods on A800 80G GPUs. V represents Vicuna, L31 represents LLaMA-Instruct 3.1, L33 represents LLaMA-Instruct 3.3, and DSL represents DeepSeek-R1-Distill-LLaMA. }
  \label{tab:a800_experiments}
\end{table*}

\begin{table*}[!ht]
  \centering
  \resizebox{\linewidth}{!}
  {\footnotesize
    \begin{tabular}{cccccccccccccc}
    \toprule
          &       & \multicolumn{2}{c}{MT-bench} & \multicolumn{2}{c}{HumanEval} & \multicolumn{2}{c}{GSM8K} & \multicolumn{2}{c}{Alpaca} & \multicolumn{2}{c}{CNN/DM} & \multicolumn{2}{c}{Mean} \\
    \midrule
    Model & Method & Speedup & $\tau$     & Speedup & $\tau$     & Speedup & $\tau$     & Speedup & $\tau$     & Speedup & $\tau$     & Speedup & $\tau$ \\
      &  & Calls & Tokens     & Calls & Tokens    & Calls & Tokens     & Calls & Tokens     & Calls & Tokens     & Calls & Tokens \\
    \midrule

\multirow{4}{*}{V 13B}
      & \multirow{2}{*}{EAGLE-3} & 2.19$\times$ & 6.53  & 2.56$\times$ & 7.74  & 2.15$\times$ & 6.39  & 2.14$\times$ & 6.54  & 1.96$\times$ & 6.54  & 2.20$\times$ & 6.75 \\
      & &  6312 &  329721 &  3428 &  171647 &  2432 &  124019 &  2077 &  109123 &  1980 &  101430 &  3246 &  167188 \\
\cmidrule(lr){2-14}
      & \multirow{2}{*}{HeteroSpec} & \textbf{2.45$\times$} & \textbf{7.07}  & \textbf{2.94$\times$} & \textbf{9.05}  & \textbf{2.31$\times$} & \textbf{6.68}  & \textbf{2.25$\times$} & \textbf{6.70}  & \textbf{2.14$\times$} & \textbf{7.05}  & \textbf{2.42$\times$} & \textbf{7.31} \\
      & &  \textbf{5922} &  \textbf{278261} &  \textbf{2988} &  \textbf{119741} &  \textbf{2327} &  \textbf{111170} &  \textbf{2012} &  \textbf{102442} &  \textbf{1877} &  \textbf{87072} &  \textbf{3025} &  \textbf{139737} \\
    \midrule

\multirow{4}{*}{L31 8B}
      & \multirow{2}{*}{EAGLE-3} & 2.67$\times$ & 6.15  & 3.06$\times$ & 6.83  & 2.85$\times$ & 6.34  & 2.97$\times$ & 6.84  & 2.20$\times$ & 5.38  & 2.75$\times$ & 6.31 \\
      & &  8687 &  541384 &  4129 &  249275 &  2446 &  149447 &  2586 &  163017 &  3996 &  244437 &  4369 &  269512 \\
\cmidrule(lr){2-14}
      & \multirow{2}{*}{HeteroSpec} & \textbf{2.79$\times$} & \textbf{6.48}  & \textbf{3.31$\times$} & \textbf{7.46}  & \textbf{2.92$\times$} & \textbf{6.54}  & \textbf{3.16$\times$} & \textbf{7.08}  & \textbf{2.31$\times$} & \textbf{5.49}  & \textbf{2.90$\times$} & \textbf{6.61} \\
      & &  \textbf{8344} &  \textbf{488135} &  \textbf{3832} &  \textbf{198173} &  \textbf{2386} &  \textbf{141866} &  \textbf{2492} &  \textbf{153535} &  \textbf{3907} &  \textbf{234268} &  \textbf{4192} &  \textbf{243195} \\
    \midrule
    
\multirow{4}{*}{DSL 8B}
      & \multirow{2}{*}{EAGLE-3} & 2.65$\times$ & 5.88  & 2.97$\times$ & 6.59  & 3.18$\times$ & 7.09  & 2.54$\times$ & 5.62  & 2.15$\times$ & 5.03  & 2.70$\times$ & 6.04 \\
      & &  6994 &  426983 &  6327 &  387276 &  4810 &  295472 &  7066 &  432824 &  8133 &  499140 &  6666 &  408339 \\
\cmidrule(lr){2-14}
      & \multirow{2}{*}{HeteroSpec} & \textbf{2.92$\times$} & \textbf{6.14}  & \textbf{3.28$\times$} & \textbf{6.83}  & \textbf{3.69$\times$} & \textbf{7.88}  & \textbf{2.79$\times$} & \textbf{5.76}  & \textbf{2.29$\times$} & \textbf{5.08}  & \textbf{2.99$\times$} & \textbf{6.34} \\
      & &  \textbf{6775} &  \textbf{396712} &  \textbf{6133} &  \textbf{353382} &  \textbf{4341} &  \textbf{218875} &  \textbf{6942} &  \textbf{416289} &  \textbf{8076} &  \textbf{484295} &  \textbf{6453} &  \textbf{373911} \\
    \bottomrule
    \end{tabular}
    }
  \caption{Speedup ratios, average acceptance lengths ($\tau$), total validation calls (Calls), and total verification tokens (Tokens) of different methods on L40 40G GPUs. V represents Vicuna, L31 represents LLaMA-Instruct 3.1, and DSL represents DeepSeek-R1-Distill-LLaMA. }
  \label{tab:l40_experiments}
\end{table*}

Since speedup ratio is a hardware-dependent metric, we conducted comprehensive experiments on A800 80G and L40 40G GPUs, beyond the H20-3e 141G GPU, to demonstrate HeteroSpec's excellent portability and hardware compatibility. As shown in Table~\ref{tab:a800_experiments} and Table~\ref{tab:l40_experiments}, HeteroSpec maintains consistent performance advantages across all testing platforms, comprehensively outperforming the state-of-the-art EAGLE-3 across key metrics, demonstrating superior hardware generalization capabilities and cross-platform stability.

\subsection{Discussion on Temperature = 1}
\label{app:t1}

\begin{table*}[!htbp]
  \centering
\resizebox{\linewidth}{!}
  {\footnotesize
    \begin{tabular}{cccccccccccccc}
    \toprule
          &       & \multicolumn{2}{c}{MT-bench} & \multicolumn{2}{c}{HumanEval} & \multicolumn{2}{c}{GSM8K} & \multicolumn{2}{c}{Alpaca} & \multicolumn{2}{c}{CNN/DM} & \multicolumn{2}{c}{Mean} \\
    \midrule
    Model & Method & Speedup & $\tau$     & Speedup & $\tau$     & Speedup & $\tau$     & Speedup & $\tau$     & Speedup & $\tau$     & Speedup & $\tau$ \\
      &  & Calls & Tokens     & Calls & Tokens    & Calls & Tokens     & Calls & Tokens     & Calls & Tokens     & Calls & Tokens \\
    \midrule

\multirow{4}{*}{V 13B}
      & \multirow{2}{*}{EAGLE-3} & 2.36$\times$ & 5.56  & 2.81$\times$ & 6.42  & 2.36$\times$ & 5.66  & 2.37$\times$ & 5.48  & 2.12$\times$ & 5.58  & 2.40$\times$ & 5.74 \\
      & &  7637 &  393862 &  4602 &  231721 &  2131 &  110544 &  2718 &  137837 &  2207 &  112553 &  3859 &  197303 \\
\cmidrule(lr){2-14}
      & \multirow{2}{*}{HeteroSpec} & \textbf{2.50$\times$} & \textbf{5.90}  & \textbf{3.08$\times$} & \textbf{7.25}  & \textbf{2.48$\times$} & \textbf{5.78}  & \textbf{2.49$\times$} & \textbf{5.78}  & \textbf{2.34$\times$} & \textbf{6.18}  & \textbf{2.58$\times$} & \textbf{6.18} \\
      & &  \textbf{7392} &  \textbf{363224} &  \textbf{4154} &  \textbf{182479} &  \textbf{1853} &  \textbf{95514} &  \textbf{2508} &  \textbf{127051} &  \textbf{1943} &  \textbf{92877} &  \textbf{3570} &  \textbf{172229} \\
    \midrule

\multirow{4}{*}{L31 8B}
      & \multirow{2}{*}{EAGLE-3} & 2.47$\times$ & 4.47  & 3.42$\times$ & 6.04  & 3.11$\times$ & 5.57  & 3.13$\times$ & 5.67  & 2.38$\times$ & 4.44  & 2.90$\times$ & 5.24 \\
      & &  14912 &  917450 &  5414 &  326152 &  3290 &  199951 &  3451 &  213993 &  4746 &  289631 &  6363 &  389435 \\
\cmidrule(lr){2-14}
      & \multirow{2}{*}{HeteroSpec} & \textbf{2.67$\times$} & \textbf{4.84}  & \textbf{3.71$\times$} & \textbf{6.61}  & \textbf{3.20$\times$} & \textbf{5.62}  & \textbf{3.26$\times$} & \textbf{5.78}  & \textbf{2.44$\times$} & \textbf{4.54}  & \textbf{3.06$\times$} & \textbf{5.48} \\
      & &  \textbf{13298} &  \textbf{820799} &  \textbf{4992} &  \textbf{265759} &  \textbf{3607} &  \textbf{213785} &  \textbf{3103} &  \textbf{190052} &  \textbf{4573} &  \textbf{278893} &  \textbf{5915} &  \textbf{353858} \\
    \midrule

\multirow{4}{*}{L33 70B}
      & \multirow{2}{*}{EAGLE-3} & 5.12$\times$ & 5.50  & 5.89$\times$ & 6.14  & 5.63$\times$ & 5.98  & 5.69$\times$ & 6.49  & 4.21$\times$ & 4.90  & 5.31$\times$ & 5.80 \\
      & &  10375 &  514086 &  5250 &  252578 &  2552 &  123939 &  2946 &  147345 &  4355 &  210372 &  5096 &  249664 \\
\cmidrule(lr){2-14}
      & \multirow{2}{*}{HeteroSpec} & \textbf{5.21$\times$} & \textbf{5.63}  & \textbf{6.15$\times$} & \textbf{6.54}  & \textbf{5.78$\times$} & \textbf{6.16}  & \textbf{5.78$\times$} & \textbf{6.68}  & \textbf{4.35$\times$} & \textbf{5.05}  & \textbf{5.45$\times$} & \textbf{6.01} \\
      & &  \textbf{9973} &  \textbf{458586} &  \textbf{4857} &  \textbf{190939} &  \textbf{2454} &  \textbf{108569} &  \textbf{2925} &  \textbf{137289} &  \textbf{4258} &  \textbf{200804} &  \textbf{4893} &  \textbf{219237} \\
    \midrule
    
\multirow{4}{*}{DSL 8B}
      & \multirow{2}{*}{EAGLE-3} & 2.77$\times$ & 4.96  & 3.07$\times$ & 5.42  & 3.82$\times$ & 6.67  & 2.65$\times$ & 4.50  & 2.31$\times$ & 4.33  & 2.92$\times$ & 5.18 \\
      & &  8408 &  519731 &  7680 &  470112 &  4861 &  298481 &  8865 &  542741 &  9280 &  570589 &  7819 &  480331 \\
\cmidrule(lr){2-14}
      & \multirow{2}{*}{HeteroSpec} & \textbf{2.98$\times$} & \textbf{5.17}  & \textbf{3.18$\times$} & \textbf{5.54}  & \textbf{4.21$\times$} & \textbf{7.53}  & \textbf{2.71$\times$} & \textbf{4.54}  & \textbf{2.39$\times$} & \textbf{4.39}  & \textbf{3.09$\times$} & \textbf{5.43} \\
      & &  \textbf{8191} &  \textbf{501969} &  \textbf{7560} &  \textbf{456374} &  \textbf{4427} &  \textbf{238975} &  \textbf{8480} &  \textbf{516930} &  \textbf{9029} &  \textbf{553813} &  \textbf{7537} &  \textbf{453612} \\
    \bottomrule
    \end{tabular}
    }
  \caption{Speedup ratios, average acceptance lengths ($\tau$), total validation calls (Calls), and total verification tokens (Tokens) of different methods on H20-3e GPUs with Temperature=1. V represents Vicuna, L31 represents LLaMA-Instruct 3.1, L33 represents LLaMA-Instruct 3.3, and DSL represents DeepSeek-R1-Distill-LLaMA. }
  \label{tab:h20_t1_experiments}
\end{table*}

For the temperature=1 setting, we adopt the same experimental configuration and conduct experiments on Vicuna 13B, LLaMA-Instruct 3.1 8B, LLaMA-Instruct 3.3 70B, and DeepSeek-R1-Distill-LLaMA. For the 8B/13B models, we employ a NVIDIA H20-3e 141G GPU. For the 70B model, we use two H20-3e GPUs due to memory limitations.

Under temperature=1, the flatter output distribution makes it significantly more challenging for the draft model to accurately predict the target model's stochastic choices, typically resulting in reduced acceleration gains from speculative decoding. Although more decoding instances fall into high-entropy bins, HeteroSpec is still capable of dynamically optimizing speculative extension and re-pruning strategies based on "decoding difficulty," thereby maximizing the accepted token length per expensive target model invocation. As demonstrated in Table~\ref{tab:h20_t1_experiments}, HeteroSpec consistently outperforms the state-of-the-art EAGLE-3 across key performance metrics, including speedup ratio, acceptance length, and verification overhead, showcasing remarkable robustness.

\subsection{More Discussion on HeteroSpec}
\label{sec:54}

HeteroSpec demonstrates strong potential for large-scale LLM service systems using speculative decoding~\cite{liu2024optimizingspeculativedecodingserving,sadhukhan2025magicdecbreakinglatencythroughputtradeoff,li2025adaserveslocustomizedllmserving}. Different applications impose diverse service-level objectives (SLOs) on inference latency: chatbots tolerate $200\sim500$ ms response times, while web search and autonomous driving require stricter constraints of $20\sim100$ ms~\cite{BRYSBAERT2019104047,zhong2024distservedisaggregatingprefilldecoding,lin2024parrotefficientservingllmbased}. SLO-customized LLM service systems are thus designed to dynamically select tokens to meet these individualized latency requirements while optimizing overall throughput.

Such problems are modeled using a hardware budget—the maximum tokens processed per forward pass~\cite{li2025adaserveslocustomizedllmserving}. Given a hardware budget and a batch of requests, the SLO-customized system aims to (1) satisfy diverse SLO requirements—typically measured by TPOT (Time Per Output Token)—and (2) maximize token acceptance during verification. HeteroSpec's Dynamic Extended Drafting increases acceptance for low-entropy requests, while Top-$N$ Pruning reduces their budget requirements without sacrificing acceptance rates. This enables budget reallocation to stricter SLO requests, improving overall SLO satisfaction. As an orthogonal strategy to existing schedulers, HeteroSpec can significantly enhance throughput in SLO-customized systems, with future integration planned for inference services.

\subsection{Future Work}
\label{app:Future}

Future work includes extending HeteroSpec to support additional orthogonal speculative decoding strategies, such as asynchronous draft-and-target, for enhanced computational efficiency through overlapping execution. We will also explore its application to ultra-long sequence generation, addressing pronounced resource and verification overheads to further demonstrate universality and scalability. Another key direction is integrating HeteroSpec into SLO-aware inference service systems. We envision it as an auxiliary module alongside existing scheduling and batching strategies. By leveraging its orthogonality, HeteroSpec can enhance system performance, provide flexible resource management, and maximize overall throughput and SLO satisfaction in real-world LLM service deployments.

\end{document}